\documentclass{article}

\usepackage{microtype}
\usepackage{graphicx}
\usepackage{subfigure}
\usepackage{booktabs} 

\usepackage{hyperref}

\usepackage[accepted]{icml2023}

\usepackage{amsmath}
\usepackage{amssymb}
\usepackage{mathtools}
\usepackage{amsthm}

\usepackage[capitalize,noabbrev]{cleveref}

\usepackage{mathrsfs}
\usepackage{multirow}
\usepackage{array}
\usepackage{bm}
\usepackage{algorithm}
\usepackage{algorithmic}
\usepackage{epstopdf}

\theoremstyle{plain}
\newtheorem{theorem}{Theorem}[section]

\newtheorem{lemma}[theorem]{Lemma}

\theoremstyle{definition}
\newtheorem{definition}[theorem]{Definition}

\theoremstyle{remark}

\usepackage[textsize=tiny]{todonotes}

\icmltitlerunning{Deep Anomaly Detection with Scale Learning}

\begin{document}
	
	\twocolumn[
	\icmltitle{	Fascinating Supervisory Signals and Where to Find Them: \\ Deep Anomaly Detection with Scale Learning}
	
	
	
	
	\begin{icmlauthorlist}
		\icmlauthor{Hongzuo Xu}{pdl,coc}
		\icmlauthor{Yijie Wang}{pdl,coc}
		\icmlauthor{Juhui Wei}{cos}
		\icmlauthor{Songlei Jian}{coc}
		\icmlauthor{Yizhou Li}{pdl,coc}
		\icmlauthor{Ning Liu}{pdl,coc}
	\end{icmlauthorlist}
	
	\icmlaffiliation{pdl}{National Key Laboratory of Parallel and Distributed Computing}
	\icmlaffiliation{coc}{College of Computer, National University of Defense Technology}
	\icmlaffiliation{cos}{College of Science, National University of Defense Technology}
	
	\icmlcorrespondingauthor{Yijie Wang}{wangyijie@nudt.edu.cn}
	
	\icmlkeywords{Anomaly Detection, Self-supervised Learning, Deep Learning, Tabular Data}
	
	\vskip 0.3in
	]
	
	
	
	\printAffiliationsAndNotice{}  
	
	\begin{abstract}

		Due to the unsupervised nature of anomaly detection, the key to fueling deep models is finding supervisory signals. Different from current reconstruction-guided generative models and transformation-based contrastive models, we devise novel data-driven supervision for tabular data by introducing a characteristic -- \textit{scale} -- as data labels. By representing varied sub-vectors of data instances, we define \textit{scale} as the relationship between the dimensionality of original sub-vectors and that of representations. Scales serve as labels attached to transformed representations, thus offering ample labeled data for neural network training. This paper further proposes a scale learning-based anomaly detection method. Supervised by the learning objective of scale distribution alignment, our approach learns the ranking of representations converted from varied subspaces of each data instance. Through this proxy task, our approach models inherent regularities and patterns within data, which well describes data ``normality". Abnormal degrees of testing instances are obtained by measuring whether they fit these learned patterns. Extensive experiments show that our approach leads to significant improvement over state-of-the-art generative/contrastive anomaly detection methods. 
		
		
	\end{abstract}

	\section{Introduction}
	
	Anomaly detection, the task of discovering exceptional data that deviate significantly from the majority \cite{aggarwal2017outlieranalysis}, has been successfully applied in many real-world domains when there is a need to identify both negative and positive rare events  (e.g., diseases, cyberspace intrusions, financial frauds, industrial faults, and marketing opportunities).
	Deep learning has shown strong modeling capability,
	which enables deep anomaly detection methods to yield drastic performance improvement over traditional methods \cite{pang2021review}.
	Due to the unsupervised nature of anomaly detection,
	\textit{
		designing deep anomaly detection models is a journey of finding reasonable supervisory signals
	}.

	Many deep anomaly detection methods \cite{xia2015learning,chen2017outlier,zhou2017anomaly,liu2019generative,liu2021rca} construct various kinds of autoencoders, generative adversarial networks, or prediction models. 
	Their learning objectives are adapted to anomaly detection by treating reconstruction errors of incoming data as abnormal degrees.
	These generative methods are intuitive and show favorable performance on several popular benchmarks. However, as has been discussed in \cite{larsen2016autoencoding,ruff2018deep,wang2022e3}, one imperfection is that their learning target is primarily designed to reconstruct/generate the whole data. That is, they are forced to focus on reducing errors in each fine-grained point, but overemphasizing low-level details may make the model hard to converge when the normal class is complicated. 
	Besides, some underlying hard anomalies can be only identified by investigating high-level pattern information in inlier data.


	To address this limitation, recent efforts \cite{golan2018deep,li2021cutpaste,wang2022e3,ristea2022self} have been made to liberate deep anomaly detection from the above reconstruction-based pipeline.
	They define various transformations and design pretext tasks, learning to classify, compare, or map these transformations. 
	They either use the learned representations for independent abnormality estimators or utilize loss values as anomaly scores.
	These models can embed high-level semantic information into the learned representations, 
	thus leading to stronger anomaly detectors with 
	promising detection accuracy.
	However, most popular transformation operations (e.g., rotations, cropping, and flip)
	can be only applied to image data.
	As for non-perceptual tabular data, it is still a non-trivial task to define suitable supervisory signals to actuate deep learning models.

	\begin{figure}[t]
		\begin{center}
			\centerline{\includegraphics[width=0.99\columnwidth]{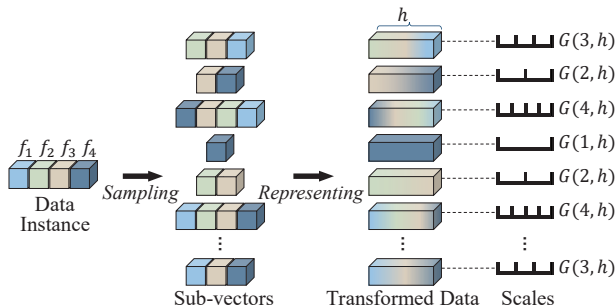}}
			\caption{
				A toy example of scales in tabular data.
				For a tabular data instance described by four features, sub-vectors with varied feature subspaces are randomly sampled from the original space and then 
				transformed into $h$-dimensional representations. 
				Scale is computed as the mathematical relationship $G$ between sub-vector length and the representation dimensionality $h$. 			
			}
			\label{fig:intro}
		\end{center}
	\end{figure}


	To this end, we introduce a new data characteristic -- \textit{scale} -- to devise a novel kind of data-driven supervision. 
	Generally, \textit{scale indicates the ratio between the real size of something and its size on a map, model, or diagram}.
	This naturally inspires us to define the scale concept in tabular data as: 
	\begin{definition}[\textit{Scale in Tabular Data}]
		Sub-vectors of tabular data instances are transformed to representations. 
		Scale is defined as the mathematical relationship between the dimensionality of sub-vectors and that of the representations, which indicates the level of detail in representations.
	\end{definition}
	\vspace{-2.8mm}
	Figure \ref{fig:intro} delineates a toy example. 
	Scales serve as labels attached to these transformed data, thereby supplying labeled data for driving neural networks on tabular data.

	However, it is still challenging to harness these labeled data.  
	Due to the feature diversity, some sub-vectors with lower dimensionality are with more details than higher-dimensional sub-vectors.
	Also, some randomly sampled subspaces might be irrelevant or even noisy.  
	These notorious problems suggest that it is unfeasible to define canonical proxy tasks like classification or simple prediction. 
	Instead, we define scale learning as follows.
	\vspace{-0.6mm}
	\begin{definition}[\textit{Scale Learning}]
		Each individual data sample in scale learning is defined as a group of representations transformed from varied sub-vectors of a data instance. 
		The predictions and corresponding labels are converted to two distributions, and scale learning is defined as a distribution alignment task.  
	\end{definition}
	\vspace{-2.6mm}
	Through optimizing distributions, the learning model is forced to focus on the relative ranking of scales rather than raw absolute values, and the increased sampling times also dilute ineffective irrelevant/noisy subspaces. 
	Our model essentially learns the listwise ranking of representations transformed from varied subspaces of each original data instance, during which our model can capture inherent regularities and patterns related to the data structure. 
	This is a novel kind of data-driven supervision that is different from current point-wise generative methods and discriminative models based on classification, comparison, or mapping.


	Based on this supervision, this paper introduces a Scale Learning-based deep Anomaly Detection method (termed SLAD).
	Concretely, SLAD first specifies a transformation function to represent sub-vectors and a labeling function to calculate data scales. After creating scale-based labeled data, SLAD performs scale learning, optimized by a distributional divergence-based loss function. 
	Through this proxy task, SLAD embeds high-level information (i.e., underlying regularities and patterns related to the data structure) into the learned scale-based ranking mechanism.
	It is similar to self-supervised learning models in vision and natural language domains that embed data semantics 
	into representations. 
	Such high-level information helps our model to tame data complexity and reveal hard anomalies.
	We extend the inlier-priority property proposed in \cite{wang2022e3} to our model. That is, due to the imbalanced nature of anomaly detection, the learning process can prioritize inliers, and the learned regularities reflect ``normality" shared in inliers.
	Anomalies, by definition, show deviated behaviors, and they cannot comply with these learned models. 
	Hence, for identifying anomalies, errors computed by the loss function are directly exploited to indicate abnormal degrees.

	Our contributions are summarized as follows:
	
	\begin{itemize}
		\setlength{\itemsep}{0pt}
		\item Conceptually, we introduce the scale concept in tabular data.
		By defining scale learning as a distribution alignment task, we appropriately harness scale-based labeled data to actuate neural networks for tabular data.
		Essentially, we devise a novel kind of data-driven supervision, and neural networks can model intrinsic regularities pertaining to the data structure. 


		\item Methodologically, we propose SLAD, a scale learning-based deep anomaly detection method. The loss values are directly exploited to indicate abnormal degrees, thus allowing SLAD to produce anomaly scores in an end-to-end fashion. 
		Our method also contributes a novel self-supervised strategy to other tasks like representation learning in tabular data. 
		
		
		\item Theoretically, we analyze the shape of the created data sample in scale learning to ensure its effectiveness in revealing anomalies. To back up the application of scale learning in anomaly detection, we examine gradient magnitude to illustrate the inlier-priority property. 
		

		\item Empirically, extensive experiments validate both the contributions in detection accuracy and the superiority in handling complicated training data. For example, SLAD raised the state-of-the-art AUC-PR from 0.82 to 0.92 (+10 points) on a popular \textit{Thyroid} benchmark.  
		

	\end{itemize}



	\section{Related work}

	
	Deep learning-empowered anomaly detection has garnered much interest recently \cite{pang2021review,ruff2020unifying}.
	Due to the unsupervised nature of anomaly detection, without readily accessible labeled training data, finding supervisory signals becomes one crucial step to fuel deep learning models for anomaly detection. This section reviews how existing studies define their learning tasks.

	One typical pipeline is based on generative methods. They take reconstruction as the learning objective to construct various kinds of autoencoders, generative adversarial networks, or prediction models and treat reconstruction errors as anomaly scores \cite{chen2017outlier,zhou2017anomaly,liu2019generative,liu2021rca,wang2021effective}. 
	Albeit intuitive, these methods overemphasize fine-grained reconstruction errors at the point-wise level, and they may fail to access high-level semantic information.

	An alternative manner is to use one-class classification to obtain a model (e.g., hypersphere or hyperplane) that can accurately describe the ``normality''.
	Many anomaly detectors \cite{ruff2018deep,goyal2020drocc,zhang2021anomaly,liznerski2021explainable} train neural networks to learn a new representation space by posing one-class constraints. 
	However, the underlying one-class assumption might be vulnerable since there is often more than one prototype in inliers. 
	Besides, some methods resort to additional label information.  
	Self-training models exploit iteratively predicted results of training data as supervisory signals while updating the model parameters \cite{pang2020self,qiu2022latent}, whereas this process might be disturbed by mislabeled data.
	It is also noteworthy that a recent method named outlier exposure introduces labeled data from other datasets, thus forming synthetically labeled data \cite{hendrycks2019oe}.

	The success of contrastive self-supervised learning in vision and natural language domains sheds light on the potential of discriminative models for embedding 
	rich semantic information into representations. 
	By using various transformation operations in image data (e.g., rotations, cropping, flip, cutout, and interpolation), many insightful approaches \cite{golan2018deep,tack2020csi,sehwag2021ssd,li2021cutpaste,wang2022e3,ristea2022self} create different views of initial data and employ classification, comparison, or mapping as pretext tasks.  
	To identify anomalies, these methods perform independent abnormality measurements upon the learned representations or directly leverage the loss function. 
	However, it is still non-trivial to define transformation operations for non-image data.

	There are very limited attempts that generalize the above contrastive strategy to tabular data. 
	GOAD \cite{bergman2020classification} is the pioneer transformation-based method that can handle non-image data, which generalizes the spatial transformation to random affine transformation. {NeuTraL} \cite{qiu2021neural} employs learnable neural transformations and proposes a noise-free, deterministic contrastive loss. 
	The literature \cite{shenkar2022internal} learns mappings that maximize the mutual information between each sampled sub-vector and the part that is masked out. 
	

	We finally review a related field, i.e., self-supervised pre-training for tabular data. 
	These models also perform contrastive learning upon different views created by corrupting random feature sub-spaces based on respective empirical marginal distribution \cite{bahri2022scarf} or feature correlations \cite{yao2021self}. 
	In addition to the corruption, the study \cite{yoon2020vime} proposes ``mask estimator'' and ``feature estimator'' heads on top of the encoder state. 

	\section{Scale Learning for Anomaly Detection}
	
	\textbf{Problem Formulation.} 
	Let $\mathcal{X}$ be the input tabular data described by a $D$-dimensional feature space.
	Each data instance $\mathbf{x} \!\in\! \mathcal{X}$ is a vector $\mathbf{x} \!\in\! \mathbb{R}^{D}$.
	By training on $\mathcal{X}$, a deep anomaly detection model builds a scoring function $\tau: \mathbb{R}^{D} \!\mapsto \! \mathbb{R}$ to quantitatively measure abnormal degrees of incoming data instances.

	
	
\begin{figure}[t]
		\centering
		\includegraphics[width=\columnwidth]{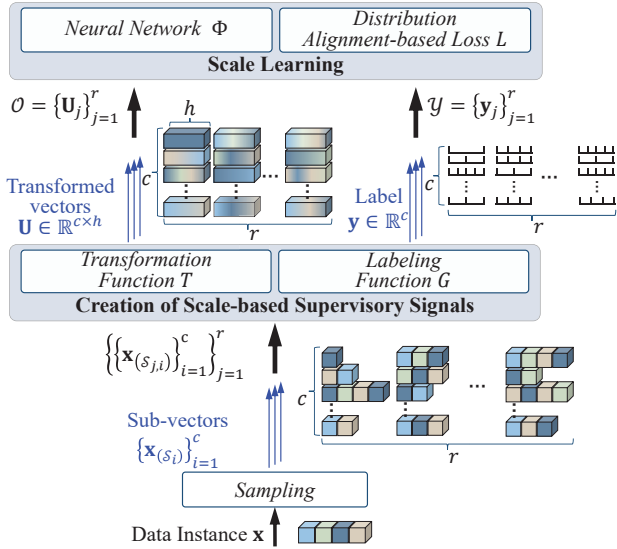}
		\caption{
			Overall framework of SLAD. For an original data instance $\mathbf{x}$, SLAD first generates a group of $c$ sub-vectors $\{\mathbf{x}_{(\mathcal{S}_i)}\}_{i=1}^{c}$ via random sampling, where $\mathbf{x}_{(\mathcal{S}_i)}$ is the sub-vector of $\mathbf{x}$ on the subspace $\mathcal{S}_{i} \subseteq \{1, \cdots, D\}$. These sub-vectors are then transformed to a unified $h$-dimensional representation frame by a \textit{Transformation function} ${T}$, yielding a matrix $\mathbf{U} \!\in\! \mathbb{R}^{c\times h}$. 
			\textit{Labeling function} ${G}$ measures scales as data labels $\mathbf{y}\! \in\! \mathbb{R}^{c}$ of transformed data in $\mathbf{U}$. 
			Each $\mathbf{U}$ and corresponding $\mathbf{y}$ are treated as one training sample, and the above process is repeated $r$ times, which produces $\mathcal{O} \in \mathbb{R}^{r\times c \times h}$ attached with labels $\mathcal{Y} \in \mathbb{R}^{r \times c} $.  
			A neural network $\Phi: \mathbb{R}^{r\times c \times h} \mapsto \mathbb{R}^{r \times c}$ is trained via the loss function $L$ to predict scale-based distributions of transformed data.
		}
		\label{fig:model}
	\end{figure}

	\textbf{Overview. }
	Figure \ref{fig:model} depicts the overall framework of SLAD. 
	We take one data instance $\mathbf{x}$ as an example to illustrate the procedure of SLAD. 
	There are two main components in SLAD. 
	\textit{The creation of scale-based supervisory signals} consists of a transformation function $T$ and a labeling function $G$, which respectively define how to transform sub-vectors of an original data instance $\mathbf{x}$ to representations $\mathbf{U}$ and how to compute scales as labels $\mathbf{y}$. 
	SLAD treats each $\mathbf{U}$ matrix that contains $c$ transformed vectors as an individual training sample, and labeled data $\mathcal{O} \!\times\! \mathcal{Y} \!=\! \{(\mathbf{U}_j, \mathbf{y}_j)\}_{j=1}^{r}$ offer supervisory signals for neural network training.
	In terms of \textit{scale learning}, we construct a neural network $\Phi$, and network parameters are optimized via a distribution alignment loss function $L$.

	\subsection{The Creation of Scale-based Supervisory Signals}
	
	\paragraph{Transformation Function $T$.}
	$T$ yields representations of sub-vectors. 
	A unified $h$-dimensional representation frame is set since these transformed data serve as training samples for downstream scale learning. $T$ can be also understood as a data preprocessing step.
	Some popular padding methods or dimensionality reduction techniques may change information contained in original sub-vectors. 
	Instead, SLAD employs neural transformation to define $T$.
	Complicated neural transformations with non-linear activation may also modify the intrinsic data structure of the input. 
	Random linear projection is a simple yet effective feature mapping technique, which can achieve dimensionality modification. 
	Thus, $T$ is defined as simple feed-forward layers that are randomly initialized, and each feature subspace corresponds to a transformation layer. 
	For a $\nu$-dimensional sub-vector $\mathbf{x}_{(\mathcal{S}_i)}\!\in\! \mathbb{R}^{\nu}$, its representation is obtained via a weight matrix $\mathbf{W}_{\nu}\in \mathbb{R}^{h \times \nu}$ and bias $b\in \mathbb{R}^{h}$, i.e.,
	\begin{equation}
		{T}(\mathbf{x}_{(\mathcal{S}_i)}) = \mathbf{W}_{\nu} \mathbf{x}_{(\mathcal{S}_i)} + b. 
	\end{equation}

	For $c$ randomly sampled sub-vectors of a data instance $\mathbf{x}$, their transformations are denoted in a matrix $\mathbf{U} \!\in\! \mathbb{R}^{c \times h}$. $\mathbf{U}$ is treated as an individual data sample for scale learning. 

	
	

	\paragraph{Labeling Function $G$.}
	SLAD further computes scales. 
	Each dimension derived via neural transformation $T$ is with equal status, so the representation dimensionality can be directly exploited. 
	However, as original tabular features are varied, we intend to weigh each feature to capture this kind of difference. 
	For ease of learning, we also increase the spacing of each scale value via a magnification factor. 
	Therefore, given the representation dimension $h$ and the feature subspace $\mathcal{S}_i$ of a sub-vector, $G$ is defined as 
	\begin{equation}
		{G}({\mathcal{S}_i}, h) = \gamma \frac{\sum_{k \in \mathcal{S}_i}  \omega_{k}}{h},
	\end{equation}
	where $\omega_k$ is the weight of the $k$th feature and $\gamma$ is a magnification factor. 
	The $\gamma$ factor is a hyper-parameter. In terms of the feature weight, a feature is more informative if it has strong interactions with other features. 
	Thus, Pearson product-moment correlation coefficient is employed. 
	Let $\mathbf{u}_{k} = \{ x_{(i, k)}\}_{i=1}^{|\mathcal{X}|} $ denote the values of the $k$th feature. 
	The weight of $k$th feature is computed as    
		$\omega_{k} \!=\! \frac{1}{ |\mathcal{F}| }\sum_{k'=1}^{|\mathcal{F}|}
		\Big\vert \frac
		{ \text{cov} (\mathbf{u}_k, \mathbf{u}_{k'})}
		{\text{dev}(\mathbf{u}_k)\text{dev}(\mathbf{u}_{k'})} \Big\vert$,   
	where $\text{cov}(\cdot,\cdot)$ and $\text{dev}(\cdot)$ denote the covariance and the standard deviation. $\omega$ ranges from 0 to 1. High-dimensional feature space is often contaminated by noisy/irrelevant features, and this calculation function might be biased.
	This function also induces considerably heavy computational overhead when handling high-dimensional data.
	Therefore, SLAD omits this step and sets $\omega\!=\!1$ for all features when the dimensionality of the original feature space is high.

	For the $\mathbf{U}$ matrix deriving from a group of sub-vectors with feature subspaces $\{\mathcal{S}_1, \cdots, \mathcal{S}_c\}$, its label is defined as a list of scales, i.e., $\mathbf{y} = \{ {G}(\mathcal{S}_1, h), \cdots, {G}(\mathcal{S}_c, h) \}$.


	\subsection{Scale Learning}
	
	The above process is repeated $r$ times for an original data instance, creating labeled data $\mathcal{O} \!\times\! \mathcal{Y} \!= \! \{(\mathbf{U}_j,\mathbf{y}_j)\}_{j\!=\!1}^{r}$. 
	SLAD constructs a neural network $\Phi: \mathbb{R}^{c\times h} \mapsto \mathbb{R}^{c}$ that maps each newly created data sample to a scale list, i.e., $\mathbf{p}\!=\!\Phi(\mathbf{U})$. 
	Scale learning is defined as a distribution alignment task to handle the feature diversity and irrelevant/noisy sampled subspaces. 
	Specifically, the predictions are processed by a softmax layer $\sigma$, i.e., $ \sigma(\mathbf{p}) \! = \! \{ \frac{\exp(p_i)}{\sum_j \exp(p_j)} \}_{i=1}^{c}$, which generates a probability distribution. $\mathbf{y}$ is also processed by a softmax function $\sigma$ to produce target distribution. 
	Listwise prediction of $c$ transformed vectors in $\mathbf{U}$ can be optimized uniformly. This way allows the optimization to be supervised by the relative values. 
	The distribution alignment task substantially teaches the network to rank representations transformed from different feature subspaces of the original data instance via the predicted scale values. 

	After obtaining the prediction $\tilde{\mathbf{p}} = \sigma(\Phi(\mathbf{U}))$ and the target $\tilde{\mathbf{y}} = \sigma(\mathbf{y})$, loss value $\ell$ is defined by a distributional divergence measure. We employ Jensen–Shannon divergence in our implementation, i.e.,
	\begin{equation}\label{eqn:ell}
		\small
		\ell(\tilde{\mathbf{p}} \Vert \tilde{\mathbf{y}}) 
		\!=\!  
		\frac{1}{2} \sum_{i=1}^{c} \tilde{p}_i \log(\frac{ \tilde{p}_i}{\frac{1}{2} (\tilde{p}_i \!+\! \tilde{y}_i) }) \!+\! 
		\frac{1}{2} \sum_{i=1}^{c} \tilde{y}_i \log(\frac{\tilde{y}_i}{\frac{1}{2} (\tilde{p}_i \!+\! \tilde{y}_i) }). \\
	\end{equation}

	The overall loss function of SLAD can be further defined as
	\begin{equation}
		L = \mathbb{E}_{\mathbf{x} \sim \mathcal{X}} \mathbb{E}_{ (\mathbf{U},\mathbf{y}) \sim \mathcal{O}_{\mathbf{x}} \times \mathcal{Y}_{\mathbf{x}}} \Big[ \ell \Big ( \sigma(\Phi(\mathbf{U})) \big\Vert \sigma(\mathbf{y}) \Big) \Big] ,
	\end{equation}
	where $\mathcal{O}_{\mathbf{x}}$ and $\mathcal{Y}_{\mathbf{x}}$ denote the supervisory signals created by an original data instance $\mathbf{x}$.


	

	\subsection{Anomaly Detection}
	
	Scale learning is not directly related to anomaly detection, and this is essentially a surrogate learning task to drive neural network training. SLAD embeds feature interactions, patterns, and inherent regularities related to the data structure into the learned scale-based ranking mechanism.
	What SLAD leverages for anomaly detection are these high-level data abstractions.

	

	We further present an \textit{inlier-priority property}. It suggests that the update of the neural network is inclined to prioritize inliers due to the imbalanced nature of anomaly detection, i.e., what SLAD derives are normal, common regularities in inliers. 
	Anomalies, by definition, are rare events and behave differently, thereby showing deviation from these learned regularities.
	This property is first proposed by \cite{wang2022e3} in which a classification-based pretext task is defined. We extend this property from the cross-entropy loss to our distributional divergence loss 
	(theoretical analysis and empirical study in Section \ref{sec:theo_inl}
	and \ref{sec:exp_closer}), which further supports the application of scale learning in anomaly detection.

	
	

	Therefore, errors computed through the loss function can indicate abnormal degrees of incoming data. For a testing data instance $\mathbf{x}$, SLAD also creates transformed data samples $\mathcal{O}$ and corresponding supervisory labels
	$\mathcal{Y}$, and its anomaly score is obtained via 
	\begin{equation}
		\mathcal{\tau(\mathbf{x})} = \sum_{(\mathbf{U}, \mathbf{y}) \in  \mathcal{O} \times \mathcal{Y}} \ell \big( \sigma(\Phi(\mathbf{U})) \big \Vert \sigma(\mathbf{y})   \big).
	\end{equation}



	\subsection{Theoretical Analysis}
	

	We explore two questions: (\textbf{Q1}) How to ensure the effectiveness of the created data sample of scale learning in revealing anomalies? (\textbf{Q2}) Does scale learning model normal regularities of inliers, thus exposing anomalies? 
	The training sample in scale learning (the $\mathbf{U}$ matrix) is generated from randomly sampled feature subspaces.
	Thus, to solve {Q1}, we consider how to determine \textit{the shape of each transformed data sample} (the sampling times $c$) such that real anomalies can still stand out in the transformed form. As for {Q2}, we examine the inlier-priority property by analyzing the gradients that determine the neural network optimization. 
	

	\paragraph{The Shape of the Data Sample of Scale Learning and its Effectiveness in Revealing Anomalies.}\label{sec:theo_size}
	
	
	
	Let $\mathbf{x}_a$ be an anomaly, and $\mathbf{U}$ indicates its transformed matrix. 
	We below derive the relationship between the size of $\mathbf{U}$ and the probability that $\mathbf{U}$ is useful to reveal $\mathbf{x}_a$ as an anomaly.  
	We assume the abnormality of $\mathbf{x}_a$ is reflected in a subspace $\mathcal{G}$ of the whole feature space $\mathcal{F}$, i.e., $\mathcal{G} \subseteq \mathcal{F}$, and $|\mathcal{G}| = \beta |\mathcal{F}|$, $\beta \in (0,1]$. The elements in $\mathcal{G}$ are effective features. 
	Generally, a subset of $\mathcal{G}$ is sufficient to discover the anomaly, and we denote this minimum size as $\alpha |\mathcal{G}|$, $\alpha \in (0,1]$. 
	Let $\mathcal{S}$ be one of the sampled subspaces when creating $\mathbf{U}$.
	The dimensionality of $\mathcal{S}$ is uniformly sampled from $1$ to $|\mathcal{F}|$.
	
	
	We first give the following Lemma (proof in Appendix \ref{appendix:probability}) to show the probability of $\mathcal{S}$ being effective.

	\begin{lemma}\label{lemma:probability}
		The probability of $\mathcal{S}$ containing at least $\alpha |\mathcal{G}|$ effective features of $\mathcal{G}$ (i.e., $\mathcal{S}$ is effective) is:
		\begin{equation}
			\small
			Pr(\mathcal{S} \text{ is useful}) \!=\! \frac{1}{|\mathcal{F}|} \!
			\sum_{j=\alpha |\mathcal{G}|}^{|\mathcal{F}|}\!
			\sum_{k=\alpha |\mathcal{G}|}^{j}\!
			\binom{j}{k} 
			\big(\!\frac{|\mathcal{G}|}{|\mathcal{F}|} \!\big)^{k} 
			\big( \!1\!-\!\frac{|\mathcal{G}|}{|\mathcal{F}|}\!\big)^{j \!-\! k}.
		\end{equation}
		
		
	\end{lemma}
	
	Based on the above Lemma, we further present an intriguing fact in the following Theorem (proof in Appendix \ref{appendix:bound}), which bounds the above probability.
	
	\begin{theorem}\label{theorem:bound}
		Given the effective feature space $\mathcal{G}$ and the minimum size $\alpha |\mathcal{G}|$ to reveal the anomaly , the lower bound of the probability of the randomly sampled subspace $\mathcal{S}$ being useful is $\inf\big(Pr(\mathcal{S} \text{ is useful}) \big) = 1 - \alpha$.
	\end{theorem}
	
	

	Let the success probability of an individual sampling be the lower bound, i.e., $1-\alpha$. We assume $\mathbf{U}$ is useful to disclose the anomaly if it has at least one element that is transformed from the effective sub-vectors. 
	Consequently, similar to Lemma \ref{lemma:probability}, the probability of $\mathbf{U}$ being useful is: 
	\begin{equation}\label{eqn:pu}
		Pr(\mathbf{U} \text{ is useful})
		= \sum_{k=1}^{c} \binom{c}{k} (1-\alpha)^k (\alpha)^{c-k}.
	\end{equation}
	$Pr(\mathbf{U} \text{ is useful})$ and $c$ are positively related, whereas a large number of useless elements in $\mathbf{U}$ may also disrupt the identification of $\mathbf{x}_a$. Thus, we use a size that is as small as possible while ensuring $Pr(\mathbf{U} \text{ is useful})$ is large enough. 
	In our default setting, we use $c\!=\!10$, which makes the probability exceed $0.999$ when $\alpha\!=\!0.5$ and $0.99$ when $\alpha\!=\!0.6$.

	\begin{table*}[htbp]
		\centering
		\caption{Detection accuracy (AUC-ROC/AUC-PR $\pm$ standard deviation) of SLAD and its competing methods. The best detector per dataset is boldfaced. ICL and GAAL run out of memory (OOM) on the ultra-high-dimensional dataset \textit{Thrombin}.    }
		\scalebox{0.73}{
			
			\begin{tabular}{ll cccccccc}
				
				\toprule
				
				& \textbf{DATA} & \textbf{SLAD} & \textbf{ICL} & \textbf{NeuTraL} & \textbf{GOAD} & \textbf{RCA} & \textbf{GAAL} & \textbf{DSVDD} & \textbf{iForest} \\
				\midrule
				
				\multirow{10}[0]{*}{\rotatebox{90}{\textbf{AUC-ROC}}} & Thyroid & \boldmath{}\textbf{0.995 $\pm$ 0.001}\unboldmath{} & 0.974 $\pm$ 0.015 & 0.985 $\pm$ 0.002 & 0.952 $\pm$ 0.005 & 0.934 $\pm$ 0.005 & 0.768 $\pm$ 0.096 & 0.930 $\pm$ 0.032 & 0.988 $\pm$ 0.002 \\
				& Arrthymia & \boldmath{}\textbf{0.825 $\pm$ 0.007}\unboldmath{} & 0.784 $\pm$ 0.048 & 0.805 $\pm$ 0.025 & 0.806 $\pm$ 0.008 & 0.767 $\pm$ 0.009 & 0.704 $\pm$ 0.082 & 0.807 $\pm$ 0.008 & 0.814 $\pm$ 0.007 \\
				& Waveform & \boldmath{}\textbf{0.812 $\pm$ 0.047}\unboldmath{} & 0.649 $\pm$ 0.048 & 0.621 $\pm$ 0.023 & 0.604 $\pm$ 0.022 & 0.626 $\pm$ 0.019 & 0.732 $\pm$ 0.074 & 0.516 $\pm$ 0.012 & 0.718 $\pm$ 0.019 \\
				& UNSW-NB15 & \boldmath{}\textbf{0.941 $\pm$ 0.004}\unboldmath{} & 0.918 $\pm$ 0.010 & 0.916 $\pm$ 0.017 & 0.903 $\pm$ 0.003 & 0.935 $\pm$ 0.001 & 0.796 $\pm$ 0.060 & 0.902 $\pm$ 0.028 & 0.758 $\pm$ 0.016 \\
				& Bank  & \boldmath{}\textbf{0.730 $\pm$ 0.004}\unboldmath{} & 0.724 $\pm$ 0.014 & 0.720 $\pm$ 0.018 & 0.587 $\pm$ 0.006 & 0.699 $\pm$ 0.003 & 0.655 $\pm$ 0.032 & 0.608 $\pm$ 0.057 & 0.723 $\pm$ 0.008 \\
				& Thrombin & \boldmath{}\textbf{0.939 $\pm$ 0.007}\unboldmath{} & OOM   & 0.460 $\pm$ 0.033 & 0.839 $\pm$ 0.011 & 0.916 $\pm$ 0.000 & OOM   & 0.520 $\pm$ 0.046 & 0.898 $\pm$ 0.008 \\
				& PageBlocks & \boldmath{}\textbf{0.972 $\pm$ 0.004}\unboldmath{} & 0.909 $\pm$ 0.025 & 0.961 $\pm$ 0.002 & 0.670 $\pm$ 0.006 & 0.864 $\pm$ 0.002 & 0.765 $\pm$ 0.032 & 0.904 $\pm$ 0.009 & 0.927 $\pm$ 0.005 \\
				& Amazon (tab) & \boldmath{}\textbf{0.605 $\pm$ 0.007}\unboldmath{} & 0.592 $\pm$ 0.005 & 0.570 $\pm$ 0.036 & 0.500 $\pm$ 0.000 & 0.538 $\pm$ 0.008 & 0.495 $\pm$ 0.032 & 0.539 $\pm$ 0.013 & 0.565 $\pm$ 0.008 \\
				& Yelp (tab)  & 0.658 $\pm$ 0.014 & \boldmath{}\textbf{0.664 $\pm$ 0.009}\unboldmath{} & 0.627 $\pm$ 0.027 & 0.501 $\pm$ 0.000 & 0.585 $\pm$ 0.008 & 0.584 $\pm$ 0.039 & 0.593 $\pm$ 0.032 & 0.609 $\pm$ 0.007 \\
				& MVTec (tab) & \boldmath{}\textbf{0.812 $\pm$ 0.009}\unboldmath{} & 0.778 $\pm$ 0.010 & 0.788 $\pm$ 0.009 & 0.666 $\pm$ 0.030 & 0.663 $\pm$ 0.022 & 0.675 $\pm$ 0.026 & 0.806 $\pm$ 0.014 & 0.757 $\pm$ 0.011 \\
				
				\midrule 
				
				\multirow{10}[0]{*}{\rotatebox{90}{\textbf{AUC-PR}}} 
				& Thyroid & \boldmath{}\textbf{0.921 $\pm$ 0.012}\unboldmath{} & 0.726 $\pm$ 0.070 & 0.824 $\pm$ 0.018 & 0.778 $\pm$ 0.008 & 0.654 $\pm$ 0.012 & 0.429 $\pm$ 0.133 & 0.470 $\pm$ 0.030 & 0.783 $\pm$ 0.037 \\
				& Arrthymia & 0.604 $\pm$ 0.006 & 0.572 $\pm$ 0.038 & 0.589 $\pm$ 0.022 & 0.631 $\pm$ 0.005 & 0.562 $\pm$ 0.009 & 0.505 $\pm$ 0.071 & 0.646 $\pm$ 0.008 & \boldmath{}\textbf{0.633 $\pm$ 0.021}\unboldmath{} \\
				& Waveform & \boldmath{}\textbf{0.432 $\pm$ 0.132}\unboldmath{} & 0.123 $\pm$ 0.040 & 0.095 $\pm$ 0.014 & 0.079 $\pm$ 0.004 & 0.088 $\pm$ 0.008 & 0.148 $\pm$ 0.060 & 0.059 $\pm$ 0.002 & 0.111 $\pm$ 0.005 \\
				& UNSW-NB15 & 0.858 $\pm$ 0.003 & \boldmath{}\textbf{0.859 $\pm$ 0.005}\unboldmath{} & 0.811 $\pm$ 0.014 & 0.813 $\pm$ 0.005 & 0.542 $\pm$ 0.009 & 0.470 $\pm$ 0.230 & 0.794 $\pm$ 0.028 & 0.111 $\pm$ 0.006 \\
				& Bank  & \boldmath{}\textbf{0.470 $\pm$ 0.003}\unboldmath{} & 0.468 $\pm$ 0.015 & 0.445 $\pm$ 0.018 & 0.300 $\pm$ 0.006 & 0.423 $\pm$ 0.002 & 0.370 $\pm$ 0.050 & 0.315 $\pm$ 0.059 & 0.449 $\pm$ 0.013 \\
				& Thrombin & \boldmath{}\textbf{0.625 $\pm$ 0.014}\unboldmath{} & OOM   & 0.038 $\pm$ 0.002 & 0.648 $\pm$ 0.013 & 0.587 $\pm$ 0.003 & OOM   & 0.074 $\pm$ 0.023 & 0.421 $\pm$ 0.017 \\
				& PageBlocks & \boldmath{}\textbf{0.872 $\pm$ 0.016}\unboldmath{} & 0.799 $\pm$ 0.033 & 0.871 $\pm$ 0.008 & 0.449 $\pm$ 0.010 & 0.739 $\pm$ 0.004 & 0.500 $\pm$ 0.034 & 0.746 $\pm$ 0.017 & 0.705 $\pm$ 0.015 \\
				& Amazon (tab) & \boldmath{}\textbf{0.120 $\pm$ 0.002}\unboldmath{} & 0.117 $\pm$ 0.001 & 0.114 $\pm$ 0.011 & 0.095 $\pm$ 0.000 & 0.105 $\pm$ 0.003 & 0.099 $\pm$ 0.008 & 0.107 $\pm$ 0.005 & 0.112 $\pm$ 0.002 \\
				& Yelp (tab)  & \boldmath{}\textbf{0.153 $\pm$ 0.005}\unboldmath{} & \boldmath{}\textbf{0.153 $\pm$ 0.003}\unboldmath{} & \boldmath{}\textbf{0.153 $\pm$ 0.015}\unboldmath{} & 0.097 $\pm$ 0.000 & 0.127 $\pm$ 0.005 & 0.125 $\pm$ 0.012 & 0.135 $\pm$ 0.013 & 0.132 $\pm$ 0.003 \\
				& MVTec (tab) & \boldmath{}\textbf{0.778 $\pm$ 0.009}\unboldmath{} & 0.740 $\pm$ 0.009 & 0.751 $\pm$ 0.011 & 0.606 $\pm$ 0.032 & 0.604 $\pm$ 0.022 & 0.618 $\pm$ 0.028 & 0.771 $\pm$ 0.017 & 0.698 $\pm$ 0.011 \\

				\bottomrule
				
			\end{tabular}%
		}
		\label{tab:effectiveness}%
	\end{table*}%

	\paragraph{Inlier-priority Property in Scale Learning.}\label{sec:theo_inl}
	
	The inlier-priority property indicates that the network optimization is inclined to prioritize inliers. 
	Since the theoretical analysis of DNNs is still intractable, we consider the same analyzable case that has been used in \cite{wang2022e3}, i.e., a feed-forward structure with sigmoid activation. 
	The penultimate layer outputs $u$ units, and the final softmax layer contains $c$ nodes. 
	Network weights are initialized by a uniform distribution $[-1,1]$.
	Considering the $k$th element ($1 \!\leq\! k \!\leq\! c$) of the prediction, the gradients w.r.t. the weights 
	(denoted as $\mathbf{w}_k = \{ w_{(s,k)} \}_{s=1}^{u}$) are directly responsible for this output. 
	Let $L_k$ be the $k$th position of the loss function of $N$ training data objects, and we can derive the expectation of gradient magnitude of updating $\mathbf{w}_k$ as follows:
	\begin{equation}\label{eqn:gradient_exp}
		\small
		\begin{split}
			\mathbb{E}\Big[ \big\Vert \nabla_{\mathbf{w}_k} L_k \big\Vert_2^2 \Big] 
			& \!=\! 
			\sum_{s=1}^{u} \mathbb{E}\Big[ \big( \sum_{i=1}^{N} \nabla_{w_{(s,k)}} \ell_k^{(i)} \big)^2 \Big] \\
			& \!=\!
			\sum_{s=1}^{u} \sum_{i=1}^{N}\sum_{j=1}^{N}
			\mathbb{E} \Big[ \nabla_{w_{(s,k)}} \ell_k^{(i)}
			\nabla_{{w}_{(s,k)}} \ell_k^{(j)}
			\Big].
		\end{split}
	\end{equation}
	$\mathbb{E}\big[\Vert \nabla_{\mathbf{w}_k} L_k \Vert_2^2 \big ]$ essentially quantifies the influence of training data on network optimization.
	We respectively denote the gradients induced by inliers and anomalies as $\nabla^{\text{inlier}}_{\mathbf{w}_k} L_k$ and $\nabla^{\text{anom}}_{\mathbf{w}_k} L_k$. Based on Taylor series expansion and gradients computation, we derive the following approximation (detailed derivation in Appendix \ref{appendix:inlier} ):
	\begin{equation}
		\small
		\frac{\mathbb{E}\big[\Vert \nabla^{\text{inlier}}_{\mathbf{w}_k} L_k \Vert_2^2 \big ]}
		{\mathbb{E}\big[\Vert \nabla^{\text{anom}}_{\mathbf{w}_k} L_k \Vert_2^2 \big ]}  
		\approx
		\frac{N_{\text{inlier}}^2}{N_{\text{anom}}^2}.
	\end{equation}
	Due to the imbalanced nature (i.e., $N_{\text{inlier}} \!\gg\! N_{\text{anom}}$), inliers govern the optimization process by inducing a significantly larger gradient magnitude. Therefore, the neural network can learn \textit{normal} regularities and patterns in inliers, thereby exposing anomalies in the inference stage. 
	


	\section{Experiments}
	
	
	\textbf{Datasets.} 
	Ten datasets are employed in our experiments. 
	\textit{Thyroid} and \textit{Arrhythmia} are two medical datasets out of four popular benchmarks used in existing studies of this research line \cite{bergman2020classification,qiu2021neural}. The other two datasets in this suite are from KDD99, while KDD99 is broadly abandoned as virtually all anomalies can be detected via one-dimensional marginal distributions. Instead, a modern intrusion detection dataset, UNSW-NB15, is exploited. Besides, our experiments also base on several datasets from different domains including \textit{Waveform} (physics), \textit{Bank} (marketing), \textit{Thrombin} (biology), and \textit{PageBlocks} (web). 
	These datasets are commonly used in anomaly detection literature \cite{pang2021review,campos2016evaluation}. 
	We employ tabular version of three vision/NLP datasets including \textit{MVTec (tab)}, \textit{Amazon (tab)}, and \textit{Yelp (tab)}, which are provided by a recent anomaly detection benchmark study \cite{han2022adbench}. 
	


	\textbf{Evaluation Protocol.}
	We follow the mainstream experimental setting of this research line \cite{bergman2020classification,qiu2021neural,shenkar2022internal} by using 50\% of normal samples for training, while the testing set contains the other half of normal samples as well as all the anomalies
	Following \cite{campos2016evaluation,pang2019deep,wang2022e3,han2022adbench,xu2019mix}, two evaluation metrics, the area under the Receiver-Operating-Characteristic curve (AUC-ROC) and the area under the Precision-Recall 
	curve (AUC-PR), are employed. 
	These two metrics can impartially evaluate the detection performance, while not posing any assumption on the anomaly threshold. 
	Unless otherwise stated, the reported metrics are averaged results over five independent runs.

	\textbf{Baseline Methods.} Seven state-of-the-art baselines are utilized. ICL \cite{shenkar2022internal}, NeuTraL \cite{qiu2021neural}, and GOAD \cite{bergman2020classification} are contrastive self-supervised methods. 
	RCA \cite{liu2021rca} and GAAL \cite{liu2019generative} are reconstruction-based generative methods.
	We also utilize DSVDD \cite{ruff2018deep}, which is a deep anomaly detection method based on one-class classification. iForest \cite{liu2008isolation} is a popular traditional (non-deep) anomaly detection baseline.

	\subsection{Anomaly Detection Performance}

	\paragraph{Effectiveness in Real-world Datasets.}\label{sec:exp_effectiveness}
	
	Table \ref{tab:effectiveness} illustrates the detection performance in terms of AUC-ROC and AUC-PR, of our model SLAD and the competing methods. 
	SLAD outperforms its state-of-the-art competing methods on eight out of ten datasets according to both two evaluation metrics. 
	SLAD averagely obtains 7\%-21\% AUC-ROC improvement and 15\%-61\% AUC-PR gain over its seven contenders. 
	Particularly, on the popular benchmark \textit{Thyroid}, SLAD raises the state-of-the-art AUC-PR by 10 points from 0.82 to 0.92.  
	We also achieve over 190\% AUC-PR leap (from 0.15 to 0.43) on \textit{Waveform}. 
	Contrastive self-supervised counterparts also show more competitive performance than reconstruction- or one-class-based methods.
	The superiority of SLAD validates the effectiveness of our scale learning task in accurately modeling normal regularities of the inherent data structure.  
	Note that SLAD performs less effectively on \textit{Arrhythmia} that has limited data instances (less than 500) since the success of neural networks generally relies on sufficient training data.
	Nevertheless, SLAD still obtains the best AUC-ROC performance on \textit{Arrhythmia}.
	
	

	\begin{figure}[t]
		\centering
		\includegraphics[width=0.99\columnwidth]{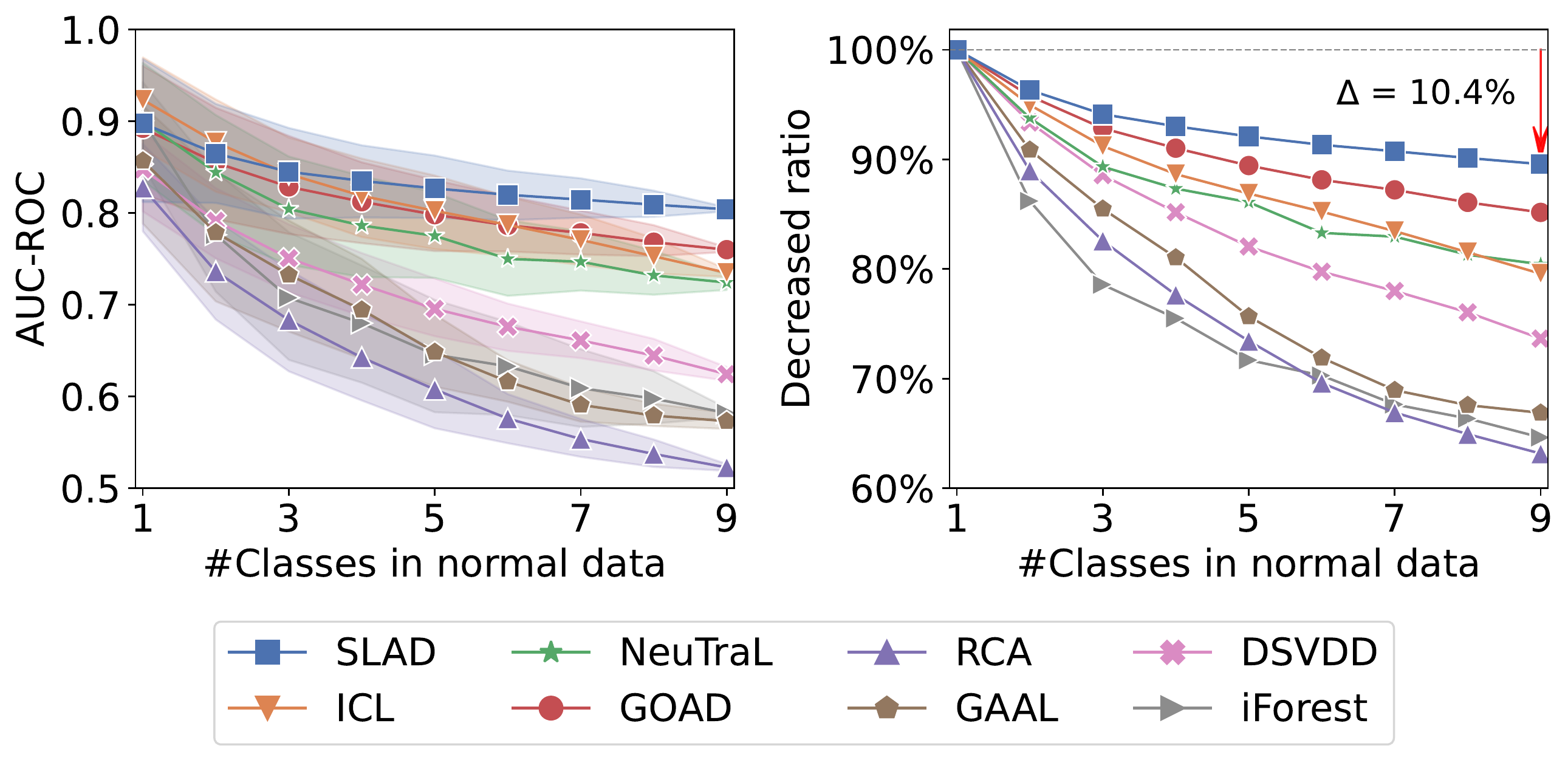}
		\caption{
			(\textit{Left}) AUC-ROC with 95\% confidence intervals on datasets with different numbers of classes considered to be inliers. (\textit{Right}) The proportion of decline compared to only one class appeared in normal data. 
		}
		\label{fig:class}
	\end{figure}

	\paragraph{Capability to Handle Complicated Normal Data.}
	
	This experiment investigates whether discriminative models can better handle complicated normal data than reconstruction- or one-class-based baselines. 
	Following \cite{qiu2021neural}, this question is empirically studied by increasing the variability of inliers. We use F-MNIST, a popular multi-class dataset, by treating each pixel as one feature. 
	A suite of datasets is created by sampling data from one class as anomalies and increasing the number of classes 
	considered to be inliers. For each case, we use nine different combinations of selected normal classes and five random seeds per class combination, producing 405 ($9\!\times\!9 \! \times \! 5$) datasets in total. 
	
	Figure \ref{fig:class} illustrates the AUC-ROC results and the decline proportion w.r.t. the increasing of class numbers in normal data. As each case corresponds to a group of datasets, we also report the 95\% confidence interval in addition to the average performance in the left panel. 
	Each detector has a comparably good performance when only one original class appears in inliers.
	The increased variety in the normal class makes the task more challenging. 
	SLAD downgrades by about 10\% and still achieves over 0.8 AUC-ROC when the normal class contains nine prototypes, while over 30\% decline is shown in generative models. 
	Reducing errors in point-wise details is hard to converge when the normal class is complicated. The one-class assumption also does not hold when there are more than two original classes considered to be inliers. 
	By contrast, SLAD, ICL, NeuTraL, and GOAD better handle the increased complexity of the normal class.
	The superiority over contrastive self-supervised counterparts validates the technical advantages of learning to rank subspace-based transformed data compared to only learning the discrimination between transformations.
	


	\begin{figure}[t]
		\centering
		\includegraphics[width=\columnwidth]{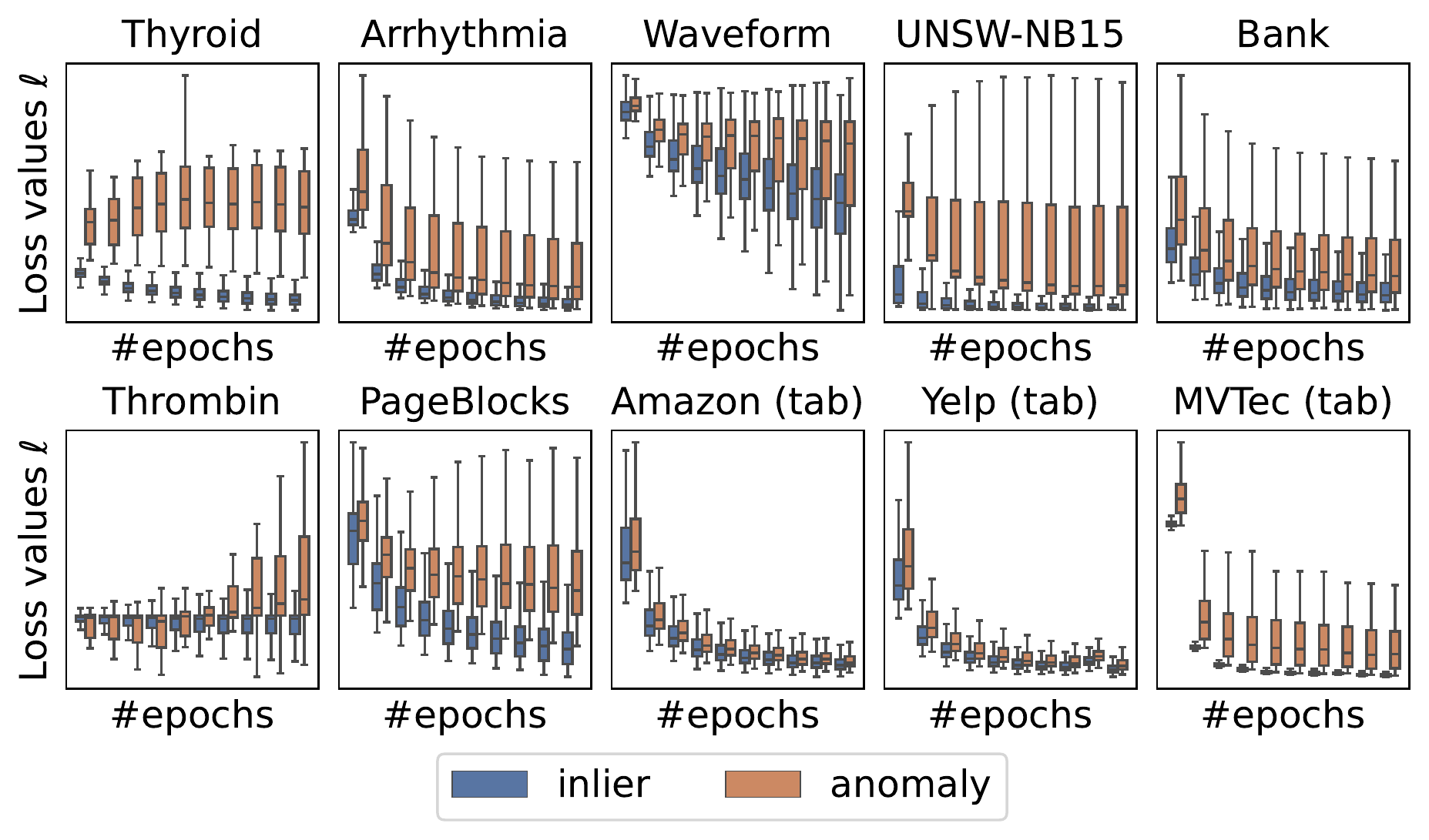}
		\vspace{-6mm}
		\caption{
			Loss values of testing inliers/anomalies during training. 
		}
		\label{fig:inlier_priority}
	\end{figure}

	\subsection{A Closer Look at Scale Learning}\label{sec:exp_closer}
	

	This experiment further investigates why scale learning can be used for anomaly detection by specifically validating the inlier-priority property and examining whether the learned regularities and patterns are class-dependent.

	\paragraph{Validating the Inlier-priority Property.}
	To look into the optimization process of scale learning, we illustrate loss values of testing inliers and anomalies per training epoch, which empirically examines the inlier-priority property. Training data are contaminated by 2\% anomalies. 
	Loss values $\ell$ of testing inliers and anomalies are respectively calculated, and Figure \ref{fig:inlier_priority} shows box plots of loss values. 
	Neural network inclines to model the inlier class, and testing inliers generally yield lower errors. 
	Compared to inliers, anomalies present significantly higher distributional divergence between derived predictions and targets, and thus two classes can be gradually separated. 
	On datasets \textit{Amazon (tab)} and \textit{Yelp (tab)}, anomalies also yield clearly reduced loss values. It might be because we employ the adapted tabular version of these two datasets, and their tabular representations are not embedded with informative features to distinguish anomalies. Other state-of-the-art competitors also fail to produce good detection results on these two datasets as shown in Table \ref{tab:effectiveness}.

	\paragraph{Validating the Class-dependency.}

\begin{figure}[t]
		\centering
		\includegraphics[width=0.99\columnwidth]{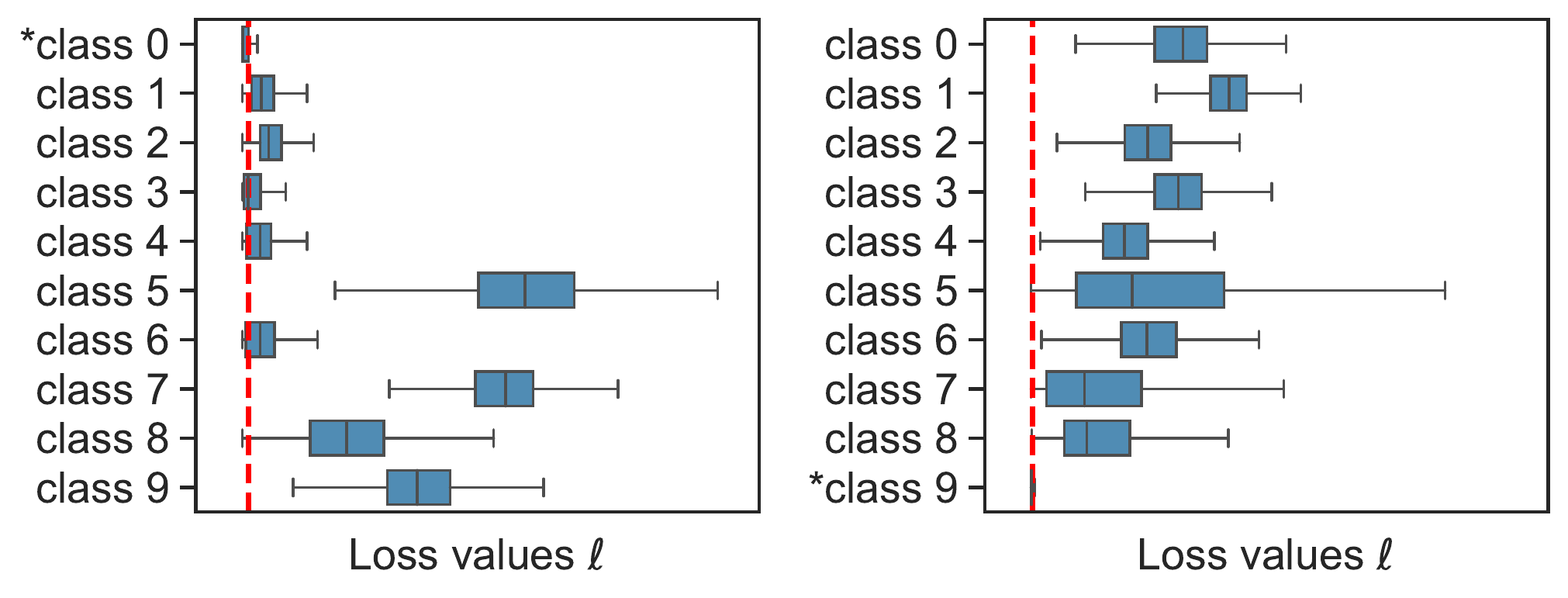}
		\vspace{-6mm}
		\caption{
			Loss values of data from the trained class and other new classes that only appear in the testing stage. 
			* indicates the trained class. 
			The red dashed line marks the upper quartile loss values of data instances from the trained class.
		}
		\label{fig:class_dependent}
	\end{figure}

	The anomaly detection performance on the used ten datasets validates that the learned regularities cannot apply to anomalies. We further delve into this question by employing the multi-class F-MNIST dataset.
	After training on one class, if data instances from new classes do not comply with the learned regularities and patterns embedded in the scale-ranking mechanism, the learned network cannot make expected predictions for data instances from new classes. 
	Therefore, this experiment tests whether the learned network generalizes to other classes by calculating loss values of data instances in new classes compared to the trained class. 
	Two cases are set by employing different original classes, i.e., class 0 (T-shirts) and class 1 (trousers), for training. 
	As shown in Figure \ref{fig:class_dependent}, loss values $\ell$ per class are denoted in box plots. The data instances from the trained class can well fit the learned network, deriving obviously lower errors. By contrast, the loss values of other classes are higher. The lower quartile of new classes is much higher than or comparable to the upper quartile of the trained class.
	These results further validate that our scale learning is class-dependent, thus further supporting its application in anomaly detection.  
	Please note that, in the left panel of Figure \ref{fig:class_dependent}, loss values in class 3 are much lower than that in other new classes since T-shirts in class 0 are semantically similar to pullovers in class 3. It is more challenging to distinguish this class. Nonetheless, data instances from this new class still show observable higher divergence than the trained class.

	\subsection{Ablation Study}
	\vspace{-1mm}
	This experiment answers two questions: (\textbf{Q1}) Can several designs in the transformation function $T$ and the labeling function $G$ be replaced with alternatives? (\textbf{Q2}) Is it necessary to define scale learning as a distribution alignment task?
	We first validate the choice of our random affine transformation function ${T}$ by replacing it with a zero padding function in \textbf{w/}~${T}_{\text{Zero}}$ and deeper feed-forward network structure in \textbf{w/}~${T}_{\text{MLP}}$, and the feature weight of the labeling function is removed in \textbf{w/o}~${G}_{\omega}$. 
	We design another three ablated versions (\textbf{w/}~${L}_{\text{ce}}$, \textbf{w/}~${L}_{\text{mse}}$, and \textbf{w/}~${L}_{\text{dcl}}$), which define scale learning as classification, regression, and contrastive learning using the cross-entropy loss, the mean-squared error, and the deterministic contrastive loss \cite{qiu2021neural}, respectively. 
	SLAD is compared with the above six ablated versions. 
	Table \ref{tab:ablation} reports the AUC-ROC results. 
	SLAD outperforms \textbf{w/}~${T}_{\text{Zero}}$ and \textbf{w/}~${T}_{\text{MLP}}$ on most of the used datasets, which validates the choice of random affine transformation when creating representations. 
	Feature weights bring an approximate 5\% improvement on the popular \textit{Thyroid} benchmark. 
	Besides, our distribution alignment-based scale learning illustrates better results than other canonical proxy tasks, which illustrates its superiority.
	It is interesting to note that \textbf{w/}~${L}_{\text{mse}}$ obtains superior average performance than \textbf{w/}~${L}_{\text{ce}}$, which implies that using clear quantitative labels may better teach the network than qualitative learning in classification.
	\textbf{w/}~${L}_{\text{dcl}}$ achieves relatively better performance since it also treats a group of transferred data as one training sample and uses the contrastive loss, i.e., it also optimizes predictions in a relative manner. 

	
	


	\begin{table}[t]
		\centering
		\caption{
			AUC-ROC performance with improvement rates of SLAD over its ablation variants per dataset. Positive rates are boldfaced. 
			\textbf{w/}~${T}_{\text{Zero}}$ cannot handle the ultra-high-dimensional data \textit{Thrombin}. 
			As SLAD only calculates feature weights on low-dimensional data,
			\textbf{w/o}~$G_{\omega}$ is performed on three datasets.
		}
		\scalebox{0.71}{
			\begin{tabular}{lclll}
				\toprule
				
				& & \multicolumn{3}{c}{\textbf{Ablation on the Creation of Supervisory Signals}} \\
				\cmidrule(lr){3-5}
				\textbf{Data} & \textbf{SLAD} & \multicolumn{1}{c}{\textbf{w/}~ $T_{\text{Zero}}$} & \multicolumn{1}{c}{\textbf{w/}~$T_{\text{MLP}}$ } & \multicolumn{1}{c}{\textbf{w/o}~${G}_{\omega}$}  \\
				
				\midrule
				
				Thyroid & 0.995 & 0.992 (\textbf{0.3\%}) & 0.995 ({0.0\%}) & 0.950 (\textbf{4.7\%}) \\
				Arrthymia & 0.825 & 0.814 (\textbf{1.4\%}) & 0.821 (\textbf{0.5\%}) & - \\
				Waveform & 0.812 & 0.759 (\textbf{7.0\%}) & 0.767 (\textbf{5.9\%}) & 0.800 (\textbf{1.5\%}) \\
				UNSW-NB15 & 0.937 & 0.933 (\textbf{0.4\%}) & 0.907 (\textbf{3.3\%}) & - \\
				Bank  & 0.730 & 0.724 (\textbf{0.8\%}) & 0.717 (\textbf{1.8\%}) & - \\
				Thrombin & 0.941 &       & 0.698 (\textbf{34.8\%}) & - \\
				PageBlocks & 0.972 & 0.971 (\textbf{0.1\%}) & 0.967 (\textbf{0.5\%}) & 0.966 (\textbf{0.6\%}) \\
				Amazon (tab) & 0.608 & 0.552 (\textbf{10.1\%}) & 0.599 (\textbf{1.5\%}) & - \\
				Yelp (tab)  & 0.661 & 0.612 (\textbf{8.0\%}) & 0.654 (\textbf{1.1\%}) & - \\
				MVTec (tab) & 0.812 & 0.775 (\textbf{4.8\%}) & 0.787 (\textbf{3.2\%}) & - \\

				\midrule
				& & \multicolumn{3}{c}{\textbf{Ablation on Scale Learning}} \\
				\cmidrule(lr){3-5}
				\textbf{Data} & \textbf{SLAD} 
				& \multicolumn{1}{c}{\textbf{w/}~${L}_{\text{ce}}$ } 
				& \multicolumn{1}{c}{\textbf{w/}~${L}_{\text{mse}}$}
				& \multicolumn{1}{c}{\textbf{w/}~${L}_{\text{dcl}}$}
				\\
				
				\midrule
				
				Thyroid & 0.995 & 0.674 (\textbf{47.6\%}) & 0.983 (\textbf{1.2\%}) & 0.978 (\textbf{1.7\%}) \\
				Arrthymia & 0.825 & 0.728 (\textbf{13.3\%}) & 0.813 (\textbf{1.5\%}) & 0.805 (\textbf{2.5\%}) \\
				Waveform & 0.812 & 0.527 (\textbf{54.1\%}) & 0.473 (\textbf{71.7\%}) & 0.770 (\textbf{5.5\%}) \\
				UNSW-NB15 & 0.937 & 0.914 (\textbf{2.5\%}) & 0.900 (\textbf{4.1\%}) & 0.922 (\textbf{1.6\%}) \\
				Bank  & 0.730 & 0.517 (\textbf{41.2\%}) & 0.732 ({-0.3\%}) & 0.714 (\textbf{2.2\%}) \\
				Thrombin & 0.941 & 0.704 (\textbf{33.7\%}) & 0.493 (\textbf{90.9\%}) & 0.626 (\textbf{50.3\%}) \\
				PageBlocks & 0.972 & 0.742 (\textbf{31.0\%}) & 0.979 ({-0.7\%}) & 0.976 ({-0.4\%}) \\
				Amazon (tab) & 0.608 & 0.536 (\textbf{13.4\%}) & 0.602 (\textbf{1.0\%}) & 0.610 ({-0.3\%}) \\
				Yelp (tab)  & 0.661 & 0.556 (\textbf{18.9\%}) & 0.666 ({-0.8\%}) & 0.676 ({-2.2\%}) \\
				MVTec (tab) & 0.812 & 0.646 (\textbf{25.7\%}) & 0.764 (\textbf{6.3\%}) & 0.776 (\textbf{4.6\%}) \\
				
				\bottomrule
			\end{tabular}
		}%
		\label{tab:ablation}%
	\end{table}%

	\section{Conclusions}
	\vspace{-1mm}
	This paper introduces SLAD, a deep anomaly detection method for tabular data.
	The core novelty of our work includes the scale concept in tabular data and a new kind of data-driven supervisory signals based on scales. 
	This supervision essentially learns the ranking of representations transformed from varied feature subspaces. It is different from current point-wise generative models and classification-, comparison-, and mapping-based discriminative models, presenting a new manner of self-supervised learning. 
	By harnessing this supervision, our model learns inherent regularities and patterns related to the data structure, which offers valuable high-level information for identifying anomalies. 
	Theoretically, we analyze how to ensure the effectiveness of the created data sample in revealing anomalies by determining its shape, and we also examine the inlier-priority property to support the application of scale learning in anomaly detection.
	Extensive experiments manifest that SLAD significantly outperforms various kinds of state-of-the-art anomaly detectors (including generative, contrastive, and one-class methods) and shows clear superiority when handling complicated data with highly varied inliers. 
	
	

	\section*{Acknowledgements}
	
	This work was supported by the National Key R\&D Program of China (No.2022ZD0115302), the National Natural Science Foundation of China (No.62002371, No.61379052), the Science Foundation of Ministry of Education of China (No.2018A02002), the Postgraduate Scientific Research Innovation Project of Hunan Province (CX20210049, CX20210028), the Natural Science Foundation for Distinguished Young Scholars of Hunan Province (No.14JJ1026), and the Foundation of National University of Defense Technology (No. ZK21-17).

	\nocite{langley00}
	
	\bibliography{main-cr}

\begin{thebibliography}{40}
\providecommand{\natexlab}[1]{#1}
\providecommand{\url}[1]{\texttt{#1}}
\expandafter\ifx\csname urlstyle\endcsname\relax
  \providecommand{\doi}[1]{doi: #1}\else
  \providecommand{\doi}{doi: \begingroup \urlstyle{rm}\Url}\fi

\bibitem[Aggarwal(2017)]{aggarwal2017outlieranalysis}
Aggarwal, C.~C.
\newblock \emph{Outlier analysis}.
\newblock Springer, 2017.
\newblock \doi{https://doi.org/10.1007/978-1-4614-6396-2}.

\bibitem[Anand et~al.(1993)Anand, Mehrotra, Mohan, and
  Ranka]{anand1993improved}
Anand, R., Mehrotra, K.~G., Mohan, C.~K., and Ranka, S.
\newblock An improved algorithm for neural network classification of imbalanced
  training sets.
\newblock \emph{IEEE Transactions on Neural Networks}, 4\penalty0 (6):\penalty0
  962--969, 1993.

\bibitem[Bahri et~al.(2022)Bahri, Jiang, Tay, and Metzler]{bahri2022scarf}
Bahri, D., Jiang, H., Tay, Y., and Metzler, D.
\newblock Scarf: Self-supervised contrastive learning using random feature
  corruption.
\newblock In \emph{International Conference on Learning Representations}, 2022.

\bibitem[Bergman \& Hoshen(2020)Bergman and Hoshen]{bergman2020classification}
Bergman, L. and Hoshen, Y.
\newblock Classification-based anomaly detection for general data.
\newblock In \emph{International Conference on Learning Representations}, 2020.

\bibitem[Campos et~al.(2016)Campos, Zimek, Sander, Campello, Micenkov{\'a},
  Schubert, Assent, and Houle]{campos2016evaluation}
Campos, G.~O., Zimek, A., Sander, J., Campello, R.~J., Micenkov{\'a}, B.,
  Schubert, E., Assent, I., and Houle, M.~E.
\newblock On the evaluation of unsupervised outlier detection: measures,
  datasets, and an empirical study.
\newblock \emph{Data mining and knowledge discovery}, 30\penalty0 (4):\penalty0
  891--927, 2016.

\bibitem[Chen et~al.(2017)Chen, Sathe, Aggarwal, and Turaga]{chen2017outlier}
Chen, J., Sathe, S., Aggarwal, C., and Turaga, D.
\newblock Outlier detection with autoencoder ensembles.
\newblock In \emph{SIAM International Conference on Data Mining}, pp.\  90--98.
  SIAM, 2017.

\bibitem[Golan \& El-Yaniv(2018)Golan and El-Yaniv]{golan2018deep}
Golan, I. and El-Yaniv, R.
\newblock Deep anomaly detection using geometric transformations.
\newblock In \emph{Advances in Neural Information Processing Systems}, pp.\
  9758--9769, 2018.

\bibitem[Goyal et~al.(2020)Goyal, Raghunathan, Jain, Simhadri, and
  Jain]{goyal2020drocc}
Goyal, S., Raghunathan, A., Jain, M., Simhadri, H.~V., and Jain, P.
\newblock {DROCC}: Deep robust one-class classification.
\newblock In \emph{Proceedings of the 37th International Conference on Machine
  Learning}, volume 119, pp.\  3711--3721. PMLR, 2020.

\bibitem[Han et~al.(2022)Han, Hu, Huang, Jiang, and Zhao]{han2022adbench}
Han, S., Hu, X., Huang, H., Jiang, M., and Zhao, Y.
\newblock Adbench: Anomaly detection benchmark.
\newblock In \emph{Advances in Neural Information Processing Systems: Datasets
  and Benchmarks Track}, 2022.

\bibitem[Hendrycks et~al.(2019)Hendrycks, Mazeika, and
  Dietterich]{hendrycks2019oe}
Hendrycks, D., Mazeika, M., and Dietterich, T.
\newblock Deep anomaly detection with outlier exposure.
\newblock In \emph{International Conference on Learning Representations}, 2019.

\bibitem[Larsen et~al.(2016)Larsen, S{\o}nderby, Larochelle, and
  Winther]{larsen2016autoencoding}
Larsen, A. B.~L., S{\o}nderby, S.~K., Larochelle, H., and Winther, O.
\newblock Autoencoding beyond pixels using a learned similarity metric.
\newblock In \emph{Proceedings of the 33rd International Conference on Machine
  Learning}, volume~48, pp.\  1558--1566. PMLR, 2016.

\bibitem[Li et~al.(2021)Li, Sohn, Yoon, and Pfister]{li2021cutpaste}
Li, C.-L., Sohn, K., Yoon, J., and Pfister, T.
\newblock Cutpaste: Self-supervised learning for anomaly detection and
  localization.
\newblock In \emph{Proceedings of the IEEE/CVF Conference on Computer Vision
  and Pattern Recognition}, pp.\  9664--9674, 2021.

\bibitem[Liu et~al.(2021)Liu, Wang, Lin, Tan, and Zhou]{liu2021rca}
Liu, B., Wang, D., Lin, K., Tan, P.-N., and Zhou, J.
\newblock Rca: A deep collaborative autoencoder approach for anomaly detection.
\newblock In \emph{Proceedings of the Thirtieth International Joint Conference
  on Artificial Intelligence}, pp.\  1505--1511, 2021.

\bibitem[Liu et~al.(2008)Liu, Ting, and Zhou]{liu2008isolation}
Liu, F.~T., Ting, K.~M., and Zhou, Z.-H.
\newblock Isolation forest.
\newblock In \emph{International Conference on Data Mining}, pp.\  413--422.
  IEEE, 2008.

\bibitem[Liu et~al.(2019)Liu, Li, Zhou, Jiang, Sun, Wang, and
  He]{liu2019generative}
Liu, Y., Li, Z., Zhou, C., Jiang, Y., Sun, J., Wang, M., and He, X.
\newblock Generative adversarial active learning for unsupervised outlier
  detection.
\newblock \emph{IEEE Transactions on Knowledge and Data Engineering},
  32\penalty0 (8):\penalty0 1517--1528, 2019.

\bibitem[Liznerski et~al.(2021)Liznerski, Ruff, Vandermeulen, Franks, Kloft,
  and Muller]{liznerski2021explainable}
Liznerski, P., Ruff, L., Vandermeulen, R.~A., Franks, B.~J., Kloft, M., and
  Muller, K.~R.
\newblock Explainable deep one-class classification.
\newblock In \emph{International Conference on Learning Representations}, 2021.

\bibitem[Pang et~al.(2019)Pang, Shen, and van~den Hengel]{pang2019deep}
Pang, G., Shen, C., and van~den Hengel, A.
\newblock Deep anomaly detection with deviation networks.
\newblock In \emph{Proceedings of the 25th ACM SIGKDD International Conference
  on Knowledge Discovery \& Data Mining}, pp.\  353--362, 2019.

\bibitem[Pang et~al.(2020)Pang, Yan, Shen, Hengel, and Bai]{pang2020self}
Pang, G., Yan, C., Shen, C., Hengel, A. v.~d., and Bai, X.
\newblock Self-trained deep ordinal regression for end-to-end video anomaly
  detection.
\newblock In \emph{Proceedings of the IEEE/CVF Conference on Computer Vision
  and Pattern Recognition}, pp.\  12173--12182, 2020.

\bibitem[Pang et~al.(2021)Pang, Shen, Cao, and Hengel]{pang2021review}
Pang, G., Shen, C., Cao, L., and Hengel, A. V.~D.
\newblock Deep learning for anomaly detection: A review.
\newblock \emph{ACM Computing Surveys}, 54\penalty0 (2), 2021.

\bibitem[Paszke et~al.(2019)Paszke, Gross, Massa, Lerer, Bradbury, Chanan,
  Killeen, Lin, Gimelshein, Antiga, et~al.]{paszke2019pytorch}
Paszke, A., Gross, S., Massa, F., Lerer, A., Bradbury, J., Chanan, G., Killeen,
  T., Lin, Z., Gimelshein, N., Antiga, L., et~al.
\newblock Pytorch: An imperative style, high-performance deep learning library.
\newblock \emph{Advances in Neural Information Processing Systems}, 32, 2019.

\bibitem[Qiu et~al.(2021)Qiu, Pfrommer, Kloft, Mandt, and
  Rudolph]{qiu2021neural}
Qiu, C., Pfrommer, T., Kloft, M., Mandt, S., and Rudolph, M.
\newblock Neural transformation learning for deep anomaly detection beyond
  images.
\newblock In \emph{Proceedings of the 38th International Conference on Machine
  Learning}, volume 139, pp.\  8703--8714. PMLR, 2021.

\bibitem[Qiu et~al.(2022)Qiu, Li, Kloft, Rudolph, and Mandt]{qiu2022latent}
Qiu, C., Li, A., Kloft, M., Rudolph, M., and Mandt, S.
\newblock Latent outlier exposure for anomaly detection with contaminated data.
\newblock In \emph{Proceedings of the 39th International Conference on Machine
  Learning}, volume 162, pp.\  18153--18167. PMLR, 2022.

\bibitem[Ristea et~al.(2022)Ristea, Madan, Ionescu, Nasrollahi, Khan, Moeslund,
  and Shah]{ristea2022self}
Ristea, N.-C., Madan, N., Ionescu, R.~T., Nasrollahi, K., Khan, F.~S.,
  Moeslund, T.~B., and Shah, M.
\newblock Self-supervised predictive convolutional attentive block for anomaly
  detection.
\newblock In \emph{Proceedings of the IEEE/CVF Conference on Computer Vision
  and Pattern Recognition}, pp.\  13576--13586, 2022.

\bibitem[Ruff et~al.(2018)Ruff, Vandermeulen, Goernitz, Deecke, Siddiqui,
  Binder, M{\"u}ller, and Kloft]{ruff2018deep}
Ruff, L., Vandermeulen, R., Goernitz, N., Deecke, L., Siddiqui, S.~A., Binder,
  A., M{\"u}ller, E., and Kloft, M.
\newblock Deep one-class classification.
\newblock In \emph{Proceedings of the 35th International Conference on Machine
  Learning}, volume~80, pp.\  4393--4402, 2018.

\bibitem[Ruff et~al.(2021)Ruff, Kauffmann, Vandermeulen, Montavon, Samek,
  Kloft, Dietterich, and Müller]{ruff2020unifying}
Ruff, L., Kauffmann, J.~R., Vandermeulen, R.~A., Montavon, G., Samek, W.,
  Kloft, M., Dietterich, T.~G., and Müller, K.-R.
\newblock A unifying review of deep and shallow anomaly detection.
\newblock \emph{Proceedings of the IEEE}, 109\penalty0 (5):\penalty0 756--795,
  2021.

\bibitem[Sehwag et~al.(2021)Sehwag, Chiang, and Mittal]{sehwag2021ssd}
Sehwag, V., Chiang, M., and Mittal, P.
\newblock Ssd: A unified framework for self-supervised outlier detection.
\newblock In \emph{International Conference on Learning Representations}, 2021.

\bibitem[Shenkar \& Wolf(2022)Shenkar and Wolf]{shenkar2022internal}
Shenkar, T. and Wolf, L.
\newblock Anomaly detection for tabular data with internal contrastive
  learning.
\newblock In \emph{International Conference on Learning Representations}, 2022.

\bibitem[Tack et~al.(2020)Tack, Mo, Jeong, and Shin]{tack2020csi}
Tack, J., Mo, S., Jeong, J., and Shin, J.
\newblock Csi: novelty detection via contrastive learning on distributionally
  shifted instances.
\newblock In \emph{Advances in Neural Information Processing Systems}, pp.\
  11839--11852, 2020.

\bibitem[Wang et~al.(2022)Wang, Zeng, Yu, Cheng, Liu, Zhou, Zhu, Kloft, Yin,
  and Liao]{wang2022e3}
Wang, S., Zeng, Y., Yu, G., Cheng, Z., Liu, X., Zhou, S., Zhu, E., Kloft, M.,
  Yin, J., and Liao, Q.
\newblock E 3 outlier: A self-supervised framework for unsupervised deep
  outlier detection.
\newblock \emph{IEEE Transactions on Pattern Analysis and Machine
  Intelligence}, 45\penalty0 (3):\penalty0 2952--2969, 2022.

\bibitem[Wang et~al.(2021)Wang, Wang, Xu, and Wang]{wang2021effective}
Wang, Z., Wang, Y., Xu, H., and Wang, Y.
\newblock Effective anomaly detection based on reinforcement learning in
  network traffic data.
\newblock In \emph{Proceedings of the IEEE 27th International Conference on
  Parallel and Distributed Systems}, pp.\  299--306. IEEE, 2021.

\bibitem[Wolpert \& Macready(1997)Wolpert and Macready]{wolpert1997no}
Wolpert, D.~H. and Macready, W.~G.
\newblock No free lunch theorems for optimization.
\newblock \emph{IEEE Transactions on Evolutionary Computation}, 1\penalty0
  (1):\penalty0 67--82, 1997.

\bibitem[Xia et~al.(2015)Xia, Cao, Wen, Hua, and Sun]{xia2015learning}
Xia, Y., Cao, X., Wen, F., Hua, G., and Sun, J.
\newblock Learning discriminative reconstructions for unsupervised outlier
  removal.
\newblock In \emph{International Conference on Computer Vision}, pp.\
  1511--1519, 2015.

\bibitem[Xu et~al.(2019)Xu, Wang, Wang, and Wu]{xu2019mix}
Xu, H., Wang, Y., Wang, Y., and Wu, Z.
\newblock Mix: A joint learning framework for detecting both clustered and
  scattered outliers in mixed-type data.
\newblock In \emph{International Conference on Data Mining}, pp.\  1408--1413.
  IEEE, 2019.

\bibitem[Xu et~al.(2021)Xu, Wang, Jian, Huang, Wang, Liu, and Li]{xu2021beyond}
Xu, H., Wang, Y., Jian, S., Huang, Z., Wang, Y., Liu, N., and Li, F.
\newblock Beyond outlier detection: Outlier interpretation by attention-guided
  triplet deviation network.
\newblock In \emph{Proceedings of the Web Conference}, pp.\  1328--1339, 2021.

\bibitem[Xu et~al.(2023)Xu, Pang, Wang, and Wang]{xu2022deep}
Xu, H., Pang, G., Wang, Y., and Wang, Y.
\newblock Deep isolation forest for anomaly detection.
\newblock \emph{IEEE Transactions on Knowledge and Data Engineering}, pp.\
  1--14, 2023.
\newblock \doi{10.1109/TKDE.2023.3270293}.

\bibitem[Yao et~al.(2021)Yao, Yi, Cheng, Yu, Chen, Menon, Hong, Chi, Tjoa,
  Kang, et~al.]{yao2021self}
Yao, T., Yi, X., Cheng, D.~Z., Yu, F., Chen, T., Menon, A., Hong, L., Chi,
  E.~H., Tjoa, S., Kang, J., et~al.
\newblock Self-supervised learning for large-scale item recommendations.
\newblock In \emph{Proceedings of the 30th ACM International Conference on
  Information \& Knowledge Management}, pp.\  4321--4330, 2021.

\bibitem[Yoon et~al.(2020)Yoon, Zhang, Jordon, and van~der
  Schaar]{yoon2020vime}
Yoon, J., Zhang, Y., Jordon, J., and van~der Schaar, M.
\newblock Vime: Extending the success of self-and semi-supervised learning to
  tabular domain.
\newblock \emph{Advances in Neural Information Processing Systems},
  33:\penalty0 11033--11043, 2020.

\bibitem[Zhang \& Deng(2021)Zhang and Deng]{zhang2021anomaly}
Zhang, Z. and Deng, X.
\newblock Anomaly detection using improved deep svdd model with data structure
  preservation.
\newblock \emph{Pattern Recognition Letters}, 148:\penalty0 1--6, 2021.

\bibitem[Zhao et~al.(2019)Zhao, Nasrullah, and Li]{zhao2019pyod}
Zhao, Y., Nasrullah, Z., and Li, Z.
\newblock Pyod: A python toolbox for scalable outlier detection.
\newblock \emph{Journal of Machine Learning Research}, 20:\penalty0 1--7, 2019.

\bibitem[Zhou \& Paffenroth(2017)Zhou and Paffenroth]{zhou2017anomaly}
Zhou, C. and Paffenroth, R.~C.
\newblock Anomaly detection with robust deep autoencoders.
\newblock In \emph{Proceedings of the 23rd ACM SIGKDD International Conference
  on Knowledge Discovery \& Data Mining}, pp.\  665--674, 2017.

\end{thebibliography}
	\bibliographystyle{icml2023}

	\newpage
	\appendix
	\onecolumn
	
	\section{Proof of Lemma \ref{lemma:probability}}\label{appendix:probability}
	
	\begin{proof}
		The subspace $\mathcal{S}$ is uniformly sampled from the full feature space $\mathcal{F}$.
		We assume the sampling process is with replacement, and the probability of sampling one effective feature is $\frac{|\mathcal{G}|}{|\mathcal{F}|}$. 
		The number of effective features in a subspace with $j$ elements is a variable that follows a binominal distribution, i.e., $X \sim b(j, \frac{|\mathcal{G}|}{|\mathcal{F}|})$, and the probability function of $X$ is:
		\begin{equation}\label{eqn:binominal}
			Pr_{j}(X\!=\!k) = 
			\binom{j}{k} (\frac{|\mathcal{G}|}{|\mathcal{F}|})^{k} (1-\frac{|\mathcal{G}|}{|\mathcal{F}|})^{j - k},
		\end{equation}

		
		The length of the sampled subspace follows a discrete uniform distribution of $\{1,\cdots, |\mathcal{F}|\}$. 
		Therefore, the probability of the subspace $\mathcal{S}$ being useful (i.e., $\mathcal{S}$ contains at least $\alpha|\mathcal{G}|$ effective features of $\mathcal{G}$) can be derived as follows.  
		
		\begin{equation}
			Pr(\mathcal{S} \text{ is useful}) = \frac{1}{|\mathcal{F}|} \sum_{j=1}^{|\mathcal{F}|} Pr_{j}(X \geq \alpha |\mathcal{G}|). 
		\end{equation}

		As $Pr_{j}(X \geq \alpha |\mathcal{G}|)$ is zero when $j < \alpha|\mathcal{G}|$, based on Equation \ref{eqn:binominal}, the probability of the sampled subspace $\mathcal{S}$ containing at least $\alpha |\mathcal{G}|$ effective features is:
		\begin{equation}
			\begin{split}
				Pr(\mathcal{S} \text{ is useful}) 
				&= \frac{1}{|\mathcal{F}|} 
				\sum_{j=\alpha |\mathcal{G}|}^{|\mathcal{F}|} 
				\sum_{k=\alpha |\mathcal{G}|}^{j}
				Pr_{j} (X=k) \\ 
				&= \frac{1}{|\mathcal{F}|} 
				\sum_{j=\alpha |\mathcal{G}|}^{|\mathcal{F}|}
				\sum_{k=\alpha |\mathcal{G}|}^{j}
				\binom{j}{k}
				(\frac{|\mathcal{G}|}{|\mathcal{F}|})^{k} (1-\frac{|\mathcal{G}|}{|\mathcal{F}|})^{j - k}
			\end{split}.
		\end{equation}
		

	\end{proof}
	
	\section{Proof of Theorem \ref{theorem:bound}}\label{appendix:bound}
	
	\begin{proof}
		
		In this proof, for the simplicity of notation, we write $|\mathcal{F}|$, $|\mathcal{G}|$, $\alpha|\mathcal{G}|$ as $n$, $m$, and $q$, repetitively, and $Pr(\mathcal{S} \text{ is useful})$ is abbreviated as $Pr$. 
		
		We first show that $Pr$ monotonically decreases with the increase of $n$. 
		\begin{equation}
			\begin{split}
				Pr_{n} - Pr_{n+1} 
				= &
				\frac{1}{n}\sum_{j=q}^{n} \sum_{k=q}^{j}
				\binom{j}{k} (\frac{m}{n})^{k} (1-\frac{m}{n})^{j - k}  
				-
				\frac{1}{n+1} \sum_{j=q}^{n+1} \sum_{k=q}^{j}
				\binom{j}{k} (\frac{m}{n+1})^{k} (1-\frac{m}{n+1})^{j - k} \\
				= & 
				\frac{1}{n}\sum_{j=q}^{n} \sum_{k=q}^{j}
				\binom{j}{k} (\frac{m}{n})^{k} (1-\frac{m}{n})^{j - k}  
				- 
				\frac{1}{n+1} \sum_{j=q}^{n} \sum_{k=q}^{j}
				\binom{j}{k} (\frac{m}{n+1})^{k} (1-\frac{m}{n+1})^{j - k} \\
				&- 
				\frac{1}{n+1} \sum_{k=q}^{n+1} \binom{n+1}{k} (\frac{m}{n+1})^{k} (1- \frac{m}{n+1})^{n+1-k} \\
				\geq &
				\frac{1}{n}\sum_{j=q}^{n} \sum_{k=q}^{j}
				\binom{j}{k} \frac{1}{n} (\frac{m}{n})^{k} (1-\frac{m}{n})^{n - k} -
				\sum_{k=q}^{n}\binom{n}{k} \frac{1}{n+1}(\frac{m}{n+1})^k (1-\frac{m}{n+1})^{n-k} \\
				&- \frac{1}{n+1} \sum_{k=q}^{n}\binom{n+1}{k}(\frac{m}{n+1})^k (1-\frac{m}{n+1})^{n+1-k} - \frac{1}{n+1}(\frac{m}{n+1})^{n+1} \\
				\geq & 
				\frac{1}{n} (\frac{m}{n})^n (1-\frac{m}{n})^{n-n} 
				- 
				\frac{1}{n+1}(\frac{m}{n+1})^n(1-\frac{m}{n+1})^{n-n} \\
				&-
				\frac{n+1}{n+1-n}\frac{1}{n+1}(\frac{m}{n+1})^n(1-\frac{m}{n+1})^{n+1-n} 
				-
				\frac{1}{n+1}(\frac{m}{n+1})^{n+1} 
				\\
				\geq & 0
			\end{split}
		\end{equation}
		
		Therefore, we have  
		\begin{equation}
			\frac{1}{n} 
			\sum_{j=q}^{n} \sum_{k=q}^{j}
			\binom{j}{k} (\frac{m}{n})^{k} (1-\frac{m}{n})^{j - k} 
			\geq 
			\lim_{n \rightarrow \infty}
			\frac{1}{n}
			\sum_{j=q}^{n} \sum_{k=q}^{j}
			\binom{j}{k} (\frac{m}{n})^{k} (1-\frac{m}{n})^{j - k}. 
		\end{equation}

		We note that 
		
		\begin{equation}\label{eqn:thm2_limeq}
			\lim_{n \rightarrow \infty}
			\frac{1}{n} 
			\sum_{j=q}^{n}\sum_{k=0}^{q-1}
			\binom{j}{k} (\frac{m}{n})^{k} (1-\frac{m}{n})^{j - k}  
			= 
			\lim_{n \rightarrow \infty}
			\sum_{j=q}^{n}\sum_{k=0}^{q-1}
			\frac{j!m^k}
			{(j-k)!k!n^{k+1}}
			(1-\frac{m}{n})^{j}.
		\end{equation}

		In addition,
		\begin{equation}
			\begin{split}
				\lim_{n \rightarrow \infty}
				\sum_{j=q}^{n}
				\frac{j!m^k}
				{(j-k)!k!n^{k+1}}
				(1-\frac{m}{n})^{j}
				&=
				\frac{1}{k!m}
				\lim_{n \rightarrow \infty}
				\sum_{j=1}^{n}
				\frac{j!}{(j-k)!}
				(\frac{m}{n})^{k+1}
				(1-\frac{m}{n})^j \\
				&=
				\frac{1}{k!m} 
				\lim_{n \rightarrow \infty}
				\sum_{j=1}^{n}
				j^k (\frac{m}{n})^{k+1}
				(1-\frac{m}{n})^j \\
				&= 
				\frac{e^{-m}}{k!m}
				\big( k!e^m - k! - k!m - m^k - {O}(m^{k-1}) \big) \\
				& \leq  \frac{1}{m}.
			\end{split}
		\end{equation}
		
		According to the above inequality, we can derive the following results from Equation (\ref{eqn:thm2_limeq}):
		\begin{equation}
			\begin{split}
				\lim_{n \rightarrow \infty}
				\frac{1}{n} 
				\sum_{j=q}^{n}\sum_{k=0}^{q-1}
				\binom{j}{k} (\frac{m}{n})^{k} (1-\frac{m}{n})^{j - k}  
				\leq 
				\sum_{k=0}^{q-1}
				\frac{1}{m}
				\leq
				\frac{q}{m}.
			\end{split}
		\end{equation}
		
		Hence, 
		\begin{equation}\label{eqn:thm2_ineq}
			\begin{split}
				\lim_{n \rightarrow \infty}
				- \frac{1}{n}
				\sum_{j=q}^{n}\sum_{k=0}^{q-1}
				\binom{j}{k}(\frac{m}{n})^{k} (1-\frac{m}{n})^{j - k} 
				&\geq 
				- \frac{q}{m} \\
				\lim_{n \rightarrow \infty}
				\frac{1}{n}
				\sum_{j=q}^{n} \Big( 1 - \sum_{k=0}^{q-1}
				\binom{j}{k} (\frac{m}{n})^{k} (1-\frac{m}{n})^{j - k} \Big) 
				&\geq  1 - \frac{q}{m} \\
				\lim_{n \rightarrow \infty}
				\frac{1}{n}
				\sum_{j=q}^{n} \sum_{k=q}^{j}
				\binom{j}{k} (\frac{m}{n})^{k} (1-\frac{m}{n})^{j - k}  
				& \geq 1 - \frac{q}{m},
			\end{split}
		\end{equation}
		
		As 
		\begin{equation}\label{eqn:thm2_decrease}
			\lim_{n \rightarrow \infty}
			\frac{1}{n}
			\sum_{j=q}^{n} \sum_{k=q}^{j}
			\binom{j}{k} (\frac{m}{n})^{k} (1-\frac{m}{n})^{j - k} 
			- 
			(1 - \frac{q}{m}) 
			= 
			\frac{e^{-m}(q! + m^{q-1} + {O}(m^{q-1}))}
			{(q-1)!m},
		\end{equation}
		i.e., Equation (\ref{eqn:thm2_decrease}) monotonically decreases when $m$ is a large number. 
		
		Also, 
		\begin{equation}\label{eqn:thm2_lim}
			\lim_{m \rightarrow \infty}
			\bigg (
			\lim_{n \rightarrow \infty}
			\frac{1}{n}
			\sum_{j=q}^{n} \sum_{k=q}^{j}
			\binom{j}{k} (\frac{m}{n})^{k} (1-\frac{m}{n})^{j - k} 
			- 
			(1 - \frac{q}{m}) 
			\bigg ) = 0 .
		\end{equation}
		
		Based on Equation (\ref{eqn:thm2_ineq})(\ref{eqn:thm2_decrease})(\ref{eqn:thm2_lim}), we finally show the lower bound of the probability is 
		\begin{equation}
			\inf (Pr)
			=  1 - \frac{q}{m} =  1- \alpha.
		\end{equation}
		
	\end{proof}

	\section{Empirical Validation of Theorem \ref{theorem:bound} }\label{appendix:bound_emp}
	
\begin{figure}[t]
		\centering
		\includegraphics[width=0.7\columnwidth]{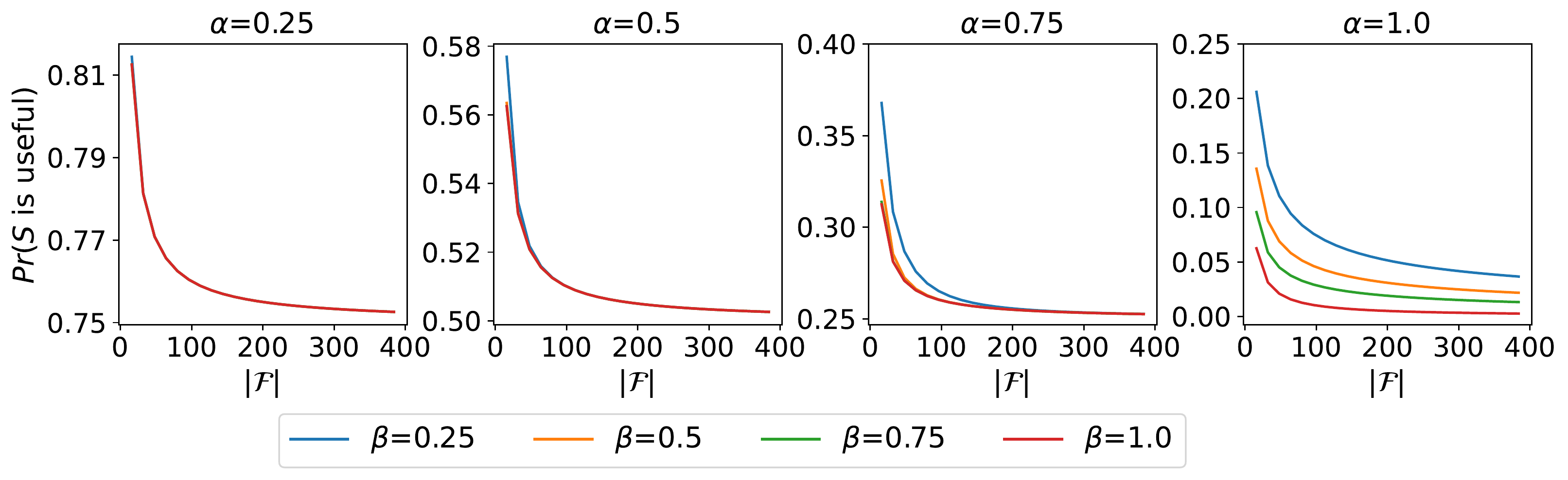}
		\caption{
			$Pr(\mathcal{S} \text{ is useful})$ of different $\alpha$ and $\beta$ settings w.r.t. the dimensionality of feature space.
		}
		\label{fig:probability}
	\end{figure}

	We further empirically inspect the lower bound of $Pr(\mathcal{S} \text{ is useful})$ in Theorem \ref{theorem:bound}. Two ratios $\alpha$ and $\beta$ are chosen from $\{0.25, 0.5, 0.75, 1.0\}$, and the data dimensionality $|\mathcal{F}|$ ranges from 0 to 400. 
	Figure \ref{fig:probability} shows how probability $Pr$ changes w.r.t. different $\alpha$, $\beta$, and $|\mathcal{F}|$.
	The probability $Pr$ monotonically decreases with the increase of $|\mathcal{F}|$, which is also proved in Appendix \ref{appendix:bound}. The function curves of $Pr$ show a clear lower bound that is determined by the $\alpha$ value. As shown in four represented $\alpha$ cases, the lower bound is shown to be $1-\alpha$, which is the same result as proved in Appendix \ref{appendix:bound}
	
	\section{Theoretical Derivation of the Inlier-Priority Property in Scale Learning }\label{appendix:inlier}
	
	We first consider the gradient $\nabla_{w_{(s,k)}} \ell_k $ in Equation (\ref{eqn:gradient_exp}). $\ell_k$ is the $k$th part of the summation in Equation (\ref{eqn:ell}), i.e.,  
	\begin{equation}
		\ell_{k} 
		= 
		\frac{1}{2} \tilde{p}_k 
		\log (\frac{\tilde{p}_k}{\frac{1}{2} (\tilde{p}_k+\tilde{y}_k)}) +
		\frac{1}{2}  \tilde{y}_k 
		\log (\frac{\tilde{y}_k}{\frac{1}{2}
			(\tilde{p}_k+\tilde{y}_k)}).
	\end{equation}  
	Thus, $\nabla_{w_{(s,k)}} \ell_k $ is given by
	\begin{equation}
		\begin{split}
			\nabla_{w_{(s,k)}} \ell_k 
			&=\! 
			\frac{\partial \ell_k }{\partial \tilde{p}_k} 
			\frac{\partial \tilde{p}_k}{\partial p_k}
			\frac{\partial p_k}{\partial w_{(s,k)}} \\  
			& =\! 
			\frac{1}{2}
			\big(\!\log 2 \!+\! \log \tilde{p}_k \!-\! \log(\tilde{y}_k \!+\! \tilde{p}_k) \big)
			\cdot
			\tilde{p}_k(1 \!-\! \tilde{p}_k) \cdot h_{s},
		\end{split}
	\end{equation}
	where $p_k$ is the prediction before the softmax layer, and $h_s$ is the output of the $s$th node in the penultimate layer. 
	
	We then consider
	$\mathbb{E} \big[ \nabla_{w_{(s,k)}} \ell_k^{(i)}
	\nabla_{{w}_{(s,k)}} \ell_k^{(j)}
	\big]$. Let $g_{i,j}^{(s,k)}(\mathbf{w}_k)\! =\! \nabla_{w_{(s,k)}} \ell_k^{(i)}
	\nabla_{{w}_{(s,k)}} \ell_k^{(j)}$. To simplify computation, we drive the following function according to the second-order Taylor series expansion:
	\begin{equation}
		g_{i,j}^{(s,k)} (\mathbf{w}_k) \approx g_{i,j}^{(s,k)} (\bm{\mu}) 
		+ \nabla_{\mathbf{w}_k} g_{i,j}^{(s,k)}(\bm{\mu}) \cdot (\mathbf{w}_k - \bm{\mu})
		+ \frac{1}{2}(\mathbf{w}_k - \bm{\mu})^{T} \cdot \nabla^2_{\mathbf{w}_k} g_{i,j}^{(s,k)}(\bm{\mu}) \cdot (\mathbf{w}_k - \bm{\mu}).
	\end{equation}
	where $\bm{\mu}$ is the expectation of $\mathbf{w}_k$. 
	
	Given that the weights are initialized by a uniform distribution $[-1,1]$, we have $\mu_{(s,k)}\!=\!0$ and $\bm{\mu} = \mathbf{0}$, 
	Hence,
	\begin{equation}
		\mathbb{E}\big[g_{i,j}^{(s,k)}(\mathbf{w}_k)\big] \approx
		\mathbb{E}\big[g_{i,j}^{(s,k)}(\mathbf{0})\big] +
		\mathbb{E}\big[\nabla_{\mathbf{w}_k} g_{i,j}^{(s,k)}(\mathbf{0}) \mathbf{w}_k \big] + 
		\mathbb{E}\big[ \frac{1}{2} \mathbf{w}_k^T \nabla^2_{\mathbf{w}_k} g_{i,j}^{(s,k)}(\mathbf{0})\mathbf{w}_k \big].
	\end{equation}

	The prediction $p_k$ is zero when network weights $\mathbf{w}_k = \mathbf{0}$, and after a softmax layer, $\tilde{p}_k=\frac{1}{c}$.  As $\tilde{y}_k$ is a constant number, we assume $\tilde{y}_k = \frac{1}{c}$ here. 
	In addition, $\mathbb{E}(w_{(s,k)}^2) = \frac{1}{3}$ and $\mathbb{E}(w_{(s,k)} w_{(t,k)}) = 0$ when $s\neq t$. Hence, we have
	\begin{equation}
		\mathbb{E}\big[g_{i,j}^{(s,k)}(\mathbf{w}_k)\big] \approx
		\frac{1}{6} \sum_{t=1}^{u}\sum_{l=1}^{c} 
		\nabla^2_{w_{t,l}} g_{i,j}^{(s,k)}(\mathbf{0}).
	\end{equation}
	
	To compute $\nabla^2_{\mathbf{w}_k} g_{i,j}^{(s,k)}(\mathbf{0})$, we first compute the following derivatives:
	\begin{equation}
		\begin{split}
			&\nabla_{w_{(t,l)}} \tilde{p}_k
			= 
			\tilde{p}_k (\delta(k,l) - \tilde{p}_k) \cdot h_t , \\
			&\nabla_{w_{(t,l)}} \tilde{p}_l
			= 
			\tilde{p}_l (1 - \tilde{p}_l) \cdot h_t , \\
			&\nabla^2_{w_{(t,l)}} \tilde{p}_k
			= 
			h_t \big( (\delta(k,l) - \tilde{p}_{l}) \cdot \nabla_{w_{(t,l)}} \tilde{p}_k - \tilde{p}_k \cdot \nabla_{w_{(t,l)}}\tilde{p}_l  \big). \\
		\end{split}
	\end{equation}
	where $\delta$ outputs whether two inputs are the same, i.e., $\delta(k,l)=1$ if $k=l$ and $\delta(k,l)=0$ otherwise. 
	
	Therefore,
	\begin{equation}
		\begin{split}
			&\nabla_{w_{(t,l)}} 
			\Big( 
			\tilde{p}_k (1-\tilde{p}_k) (\log 2 + \log \tilde{p}_k - \log(\frac{1}{c} + \tilde{p}_k ))
			\Big) \\
			= 
			&\Big( (1 - 2 \tilde{p}_k) \big(\log 2 + \log \tilde{p}_k - \log(\frac{1}{c} + \tilde{p}_k) \big) + \frac{1 - \tilde{p}_k}{1 + c \cdot \tilde{p}_k} \Big) 
			\nabla_{w_{(t,l)}}\tilde{p}_k ,
		\end{split}
	\end{equation}
	and
	\begin{equation}
		\begin{split}
			&\nabla^2_{w_{(t,l)}} \Big( 
			\tilde{p}_k (1-\tilde{p}_k) (\log 2 + \log \tilde{p}_k - \log(\frac{1}{c} + \tilde{p}_k ))
			\Big) \\
			= & \nabla_{w_{(t,l)}} 
			\bigg (\Big( (1 - 2 \tilde{p}_k) \big(\log 2 + \log \tilde{p}_k - \log(\frac{1}{c} + \tilde{p}_k) \big) + \frac{1 - \tilde{p}_k}{1 + c \cdot \tilde{p}_k} \Big) 
			\nabla_{w_{(t,l)}}\tilde{p}_k \bigg) \\
			= & -2 \big(\log2 + \log (\tilde{p}_k) - \log(\frac{1}{c} + \tilde{p}_k) \big) 
			\big(\nabla_{w_{(t,l)}}\tilde{p}_k \big)^2 + 
			(1 - 2 \tilde{p}_k) (\frac{1}{1 + c \cdot \tilde{p}_k}) 
			\big(\nabla_{w_{(t,l)}}\tilde{p}_k \big)^2  \\
			& - \frac{
				\big(1 \!+\! c \!+\! c \cdot \tilde{p}_k \big) 
				\big(\nabla_{w_{(t,l)}}\tilde{p}_k \big)^2 
				- c \cdot \big(\nabla_{w_{(t,l)}}\tilde{p}_k \big)^3
			}
			{ (1 + c \cdot \tilde{p}_k)^2 } 
			+ 
			\Big(
			(1 \!-\! 2 \tilde{p}_k) \big(\log2 \!+\! \log (\tilde{p}_k) \!-\! \log(\frac{1}{c} \!+\! \tilde{p}_k) \big) + \frac{1 - \tilde{p}_k}{1 \!+\! c \cdot \tilde{p}_k}
			\Big) 
			\nabla^2_{w_{(t,l)}}\tilde{p}_k.
		\end{split}
	\end{equation}
	If $k=l$ and $\mathbf{w}=0$, we have
	\begin{equation}
		\begin{split}
			&\nabla_{w_{(t,l)}} \Big( 
			\tilde{p}_k (1-\tilde{p}_k) (\log 2 + \log \tilde{p}_k - \log(\frac{1}{c} + \tilde{p}_k ))
			\Big)  = 
			\frac{(1-c)^2}{2 c^3} h_t ,
			\\
			&\nabla^2_{w_{(t,l)}} \Big( 
			\tilde{p}_k (1-\tilde{p}_k) (\log 2 + \log \tilde{p}_k - \log(\frac{1}{c} + \tilde{p}_k )) 
			\Big)  = 
			\frac{3 (c-3)(c-1)^2}{4 c^4} (h_t)^2.
		\end{split}
	\end{equation}
	If $k\neq l$ and $\mathbf{w}=0$, we have
	\begin{equation}
		\begin{split}
			&\nabla_{w_{(t,l)}} \Big( 
			\tilde{p}_k (1-\tilde{p}_k) (\log 2 + \log \tilde{p}_k - \log(\frac{1}{c} + \tilde{p}_k ))
			\Big)  = 
			\frac{1-c}{2 c^3} h_t ,
			\\
			&\nabla^2_{w_{(t,l)}} \Big( 
			\tilde{p}_k (1-\tilde{p}_k) (\log 2 + \log \tilde{p}_k - \log(\frac{1}{c} + \tilde{p}_k )) 
			\Big)  = 
			\frac{-9 + 7c - 2 c^2}{4 c^4} (h_t)^2.
		\end{split}
	\end{equation}
	
	Hence,
	\begin{equation}
		\begin{split}
			&\frac{1}{6} \sum_{t=1}^{u}\sum_{l=1}^{c} 
			\nabla^2_{w_{t,l}} g_{i,j}^{(s,k)}(\mathbf{0}) \\
			= & 
			\frac{1}{6} \frac{1}{4} 
			h_s^{(i)} h_s^{(j)} 
			\sum_{t=1}^{u} \big( \frac{(c^2 - c + 1)(c-1)^3}{ 4 c^6} h_t^{(i)} h_t^{(j)} \big) + 
			\frac{(c-1) (3c^3 - 14 c^2 + 16 c - 9 )}{4 c^4} 
			( (h_t^{(i)})^2 + (h_t^{(j)})^2 )).
		\end{split}
	\end{equation}
	
	The literature \cite{anand1993improved} has proved that the randomly initialized network have:
	$\mathbb{E}\big[h_s^{(i)} h_s^{(j)} \big ] \approx \frac{1}{4}$, 
	$\mathbb{E}\big[(h_s^{(i)})^2 (h_s^{(j)})^2 \big ] \approx \frac{1}{16}$, and 
	$\mathbb{E}\big[(h_s^{(i)})^3 (h_s^{(j)}) \big ] \approx \frac{1}{16}$.
	
	Therefore, 
	\begin{equation}
		\begin{split}
			\mathbb{E}\big[\Vert \nabla_{\mathbf{w}_k} L_k \Vert_2^2 \big ] 
			&\approx 
			N^2 u 
			\big( 
			\frac{(c^3 + 1)(c-1)^3}{1536 c^6 (c+1)} +
			\frac{(c-1)(3c^3 - 14 c^2 + 16c - 9)}{768 c^4} 
			\big)  \\
			& \triangleq Q N^2 ,
		\end{split}
	\end{equation}
	where $Q$ is a constant related to $c$. The gradient magnitude is proportional to the size of training data.

	We can then compare the magnitude induced by inliers and anomalies, i.e.,
	\begin{equation}
		\begin{split}
			\frac{\mathbb{E}\big[\Vert \nabla^{\text{inlier}}_{\mathbf{w}_k} L_k \Vert_2^2 \big ]}
			{\mathbb{E}\big[\Vert \nabla^{\text{anom}}_{\mathbf{w}_k} L_k \Vert_2^2 \big ]}  
			\approx
			\frac{N_{\text{inlier}}}{N_{\text{anom}}}. 
		\end{split}
	\end{equation}

	\section{Algorithms Details}
	
	We outline the detailed training procedure of SLAD in Algorithm \ref{alg:train}. SLAD uses interactions between features to calculate feature weights for the subsequent labeling function in Steps 4-5. This process is omitted by using a threshold $\delta$ in Steps 6-7 when handling high-dimensional data due to the huge computational overhead and the inaccuracy caused by irrelevant/noisy features. SLAD creates scale-based supervisory signals via the transformation function $T$ and labeling function $G$ in Steps 9-20. Every $c$ transformed vectors are contained in a $\mathbf{U}$ matrix in Steps 14-16, and finally creates $r$ matrices in $\mathcal{O}$ attached with labels in $\mathcal{Y}$ via Step 18. 
	SLAD further trains a neural network $\Phi$ to rank scale values via the loss values $\ell$ and the loss function $L$ in Steps 21-28.

	\begin{algorithm}[htbp]
		\caption{Training Procedure of SLAD}
		\label{alg:train}
		\begin{algorithmic}[1]
			\STATE {\bfseries Input:} Dataset $\mathcal{X}$.
			\STATE {\bfseries Output:} Trained neural network $\Phi^*$
			\STATE Initialize $\omega \in \mathbb{R}^{D}$ as the a feature weight list.
			\IF{$D<\delta$}
			\STATE $\omega \leftarrow  \Big \{ \frac{1}{ |\mathcal{F}| }\sum_{k'=1}^{|\mathcal{F}|}
			\big\vert \frac
			{ \text{cov} (\mathbf{u}_k, \mathbf{u}_{k'})}
			{\text{dev}(\mathbf{u}_k)\text{dev}(\mathbf{u}_{k'})} \big\vert  \Big\}_{k=1}^{D} $
			\ELSE
			\STATE $\omega \leftarrow \mathbf{1}$
			\ENDIF
			\STATE Initialize $\mathcal{O}\leftarrow \{\}$, $\mathcal{Y} \leftarrow \{\}$
			\FOR{$j=1$ {\bfseries to} $r$}
			\FOR{$\mathbf{x} \in \mathcal{X}$}
			\STATE Initialize a $c\times h$ matrix $\mathbf{U}$ for transferred data.
			\STATE Initialize a $c$-dimensional vector $\mathbf{y}$ for scale-based labels.
			\FOR{$i=1$ {\bfseries to} $c$}
			\STATE Sample a feature subset from $\mathcal{S}_i \subseteq \mathcal{F}$.
			\STATE $\mathbf{U}_{(i, \cdot)} \leftarrow {T}(\mathbf{x}_{(\mathcal{S}_i)})$, $\mathbf{y}_{(i)} \leftarrow {G}(\mathcal{S}_i, h)$
			\ENDFOR
			\STATE $\mathcal{O} \leftarrow \mathcal{O} \cup \mathbf{U}$, $\mathcal{Y} \leftarrow \mathcal{Y} \cup \mathbf{y}$
			\ENDFOR
			\ENDFOR
			\STATE Initialize parameters in neural network $\Phi$.
			\REPEAT
			\REPEAT
			\STATE Sample mini-batch training data $\mathcal{B}_x \sim \mathcal{O}$, $\mathcal{B}_y \sim \mathcal{Y}$.
			\STATE ${L} \leftarrow \mathbb{E}_{\mathbf{U} \sim \mathcal{B}_x, \mathbf{y}\sim \mathcal{B}_y} \Big [ \ell\big( \sigma(\Phi(\mathbf{U})) \Vert \sigma(\mathbf{y})  \big) \Big]$ 
			\STATE Update network parameters according to ${L}$
			\UNTIL{Reach maximum number of mini-batches}
			\UNTIL{Reach maximum training epochs}
			\STATE {\bfseries return:} $\Phi^*$
		\end{algorithmic}
	\end{algorithm}

	\section{Datasets}\label{appendix:datasets}
	Table \ref{tab:datainfo} reports the domains of the used datasets and their statistical information including data size, dimensionality, the number of anomalies (\#anom), and anomaly percentage to the whole data size (ratio). 
	These datasets are publicly available
	and they are broadly used in related literature \cite{han2022adbench,shenkar2022internal,bergman2020classification,qiu2021neural,xu2022deep}. In \textit{UNSW-NB15}, we use the network traffic of ``DoS'' attack as anomalies.
	

	\begin{table}[htbp]
		\centering
		\caption{Dataset information. Data size and dimensionality indicate the number of data instances, and the number of features. \#anom and ratio denote the number of anomalies and the abnormal percentage over all the data instances. The statistical information of MVTec (tab) is the average value over its fifteen sub-datasets.    }
		\scalebox{0.9}{
			\begin{tabular}{llllll}
				\toprule
				\textbf{Dataset} & \textbf{Domain} & \textbf{Data size}     & \textbf{Dimensionality}    & \textbf{\#anom} & \textbf{ratio} \\
				\midrule
				Thyroid & Healthcare & 3,772  & 6     & 93    & 2.5\% \\
				Arrhythmia & Healthcare & 452   & 274   & 66    & 14.6\% \\
				Waveform & Physics & 3,443  & 21    & 100   & 2.9\% \\
				UNSW-NB15 & Intrusion detection & 96,000 & 196   & 3,000  & 3.1\% \\
				Bank  & Marketing & 41,188 & 62    & 4,640  & 11.3\% \\
				Thrombin & Biology & 1,909  & 139,351 & 42    & 2.2\% \\
				PageBlocks & Web   & 5,393  & 10    & 510   & 9.5\% \\
				Amazon (tab) & NLP   & 10,000 & 768   & 500   & 5.0\% \\
				Yelp (tab)  & NLP   & 10,000 & 768   & 500   & 5.0\% \\
				MVTec (tab) & CV    & 357   & 512   & 84    & 23.5\% \\
				\midrule
		\end{tabular}}%
		\label{tab:datainfo}%
	\end{table}%

	\section{Implementation Details}

	For SLAD, we use $h\!=\!128$ in the transformation function to represent sub-vectors to a unified 128-dimensional frame. 
	Our transformation function uses $c$ and $r$ to respectively control the number of transferred data in each training sample and the repeat times. We use $c\!=\!10$ and $r\!=\!20$ by default. 
	In the labeling function $G$, we use $\delta\!=\!50$ as the threshold to omit the correlation-based weighting function. 
	The magnification factor $\gamma$ in $G$ is set as $200$ by default.
	All the deep anomaly detectors use a 1e-3 learning rate, and the mini-batch size is 128. Multi-layer perceptron network is exploited for these deep detectors except GOAD which uses the default convolutional network. We use the LeakyReLU function as the non-linear activation layer, and the hidden layer contains 100 neural units. 
	For ICL, NeuTraL, GOAD, and RCA, the representation dimensionality is set as 128, which is the same as $h$ in SLAD. 
	These datasets are split into three groups, i.e., datasets with smaller sizes including \textit{Thyroid}, \textit{Arrhythmia}, and \textit{Waveform}, transferred CV dataset \textit{MVTec (tab)}, and other datasets. We vary the number of training epochs from \{10, 20, 50, 100\} for three categories and report empirically good results.

	Our experiments are conducted on a workstation with Intel Xeon Silver 4210R CPU, a single NVIDIA TITAN RTX GPU, and 64 GB RAM.
	All the anomaly detection methods used in our experiments are implemented in Python. Deep detectors (SLAD, ICL, NeuTraL, GOAD, RCA) use the PyTorch framework \cite{paszke2019pytorch}. We implement them in the DeepOD package (\url{https://github.com/xuhongzuo/deepod}). 
	The implementation of GAAL is taken from the PyOD package \cite{zhao2019pyod}, and the non-deep baseline iForest is implemented in the Scikit-learn package. The source code of SLAD is available at \url{https://github.com/xuhongzuo/scale-learning}.

	\section{Additional Empirical Results}

	\subsection{Effectiveness on ADBench}\label{appendix:adbench}
	
	We further employ ADBench \cite{han2022adbench}, the latest anomaly detection benchmark collection of tabular data, which consists of 47 classical tabular datasets.
	These datasets are from various domains and contain varied sizes, dimensionality, anomaly types, and noise levels. It is very challenging to be the universal winner in all cases, as has been suggested in the no-free-lunch theorem \cite{wolpert1997no}. Therefore, we focus on the overall performance on this large-scale benchmark. We employ the average rank and the number/percentage of cases that each anomaly detector ranks within the Top2/Top4 positions among 8 anomaly detector candidates (our model and seven contenders). 
	Table \ref{tab:adbench} shows the experiment results. Our method SLAD successfully outperforms all of its seven state-of-the-art competing methods according to six evaluation metrics.
	It is also noteworthy that iForest is a powerful baseline, even if its workflow is very simple. In addition, the advanced self-supervised methods ICL and NeuTraL show competitive AUC-PR performance. Nevertheless, our method still leads to new state-of-the-art detection accuracy across this comprehensive benchmark collection.


	\begin{table}[t]
		\centering
		\caption{Detection performance on a benchmark collection with 47 datasets. Avg. Rank indicates the average ranking of all the datasets. \#Top2 and \#Top4 respectively count the times that each anomaly detector rank within the Top2 and Top4 positions, and \%Top2 and \%Top4 denote the corresponding percentage. 
			$\uparrow$ indicates that the higher the indicator, the better the detection performance, while $\downarrow$ denotes the lower the better.   }
		\scalebox{0.77}{
			\begin{tabular}{lcccccccccc}
				\toprule
				\multicolumn{1}{l}{\multirow{2}[0]{*}{\textbf{Model}}} & \multicolumn{5}{c}{\textbf{AUC-ROC}} & \multicolumn{5}{c}{\textbf{AUC-PR}} \\
				\cmidrule(lr){2-6}  \cmidrule(lr){7-11}
				& Avg. Rank ($\downarrow$) & \#Top2 ($\uparrow$) & \%Top2 ($\uparrow$) & \#Top4 ($\uparrow$)  & \%Top4 ($\uparrow$)  & Avg. Rank ($\downarrow$) & \#Top2 ($\uparrow$) & \%Top2 ($\uparrow$)
				& \#Top4 ($\uparrow$) & \%Top4 ($\uparrow$)
				\\
				\midrule
				iForest & 3.77  & 16    & 34.0\% & 27    & 57.4\% & 4.26  & 12    & 25.5\% & 24    & 51.1\% \\
				DSVDD & 4.66  & 10    & 21.3\% & 20    & 42.6\% & 4.70  & 12    & 25.5\% & 19    & 40.4\% \\
				GAAL  & 6.23  & 4     & 8.5\% & 9     & 19.1\% & 6.21  & 4     & 8.5\% & 8     & 17.0\% \\
				RCA   & 4.13  & 13    & 27.7\% & 26    & 55.3\% & 4.21  & 11    & 23.4\% & 26    & 55.3\% \\
				GOAD  & 6.15  & 5     & 10.6\% & 10    & 21.3\% & 5.85  & 5     & 10.6\% & 10    & 21.3\% \\
				NeuTraL & 3.68  & 16    & 34.0\% & 28    & 59.6\% & 3.77  & 17    & 36.2\% & 31    & 66.0\% \\
				ICL   & 4.09  & 12    & 25.5\% & 27    & 57.4\% & 3.64  & 16    & 34.0\% & 33    & 70.2\% \\
				SLAD (ours) & \textbf{2.98 } & \textbf{22} & \textbf{46.8\%} & \textbf{39} & \textbf{83.0\%} & \textbf{2.98 } & \textbf{21} & \textbf{44.7\%} & \textbf{38} & \textbf{80.9\%} \\

				\bottomrule
		\end{tabular}}%
		\label{tab:adbench}%
	\end{table}%
	
	\begin{figure}[t]
		\centering
		\includegraphics[width=0.9\textwidth]{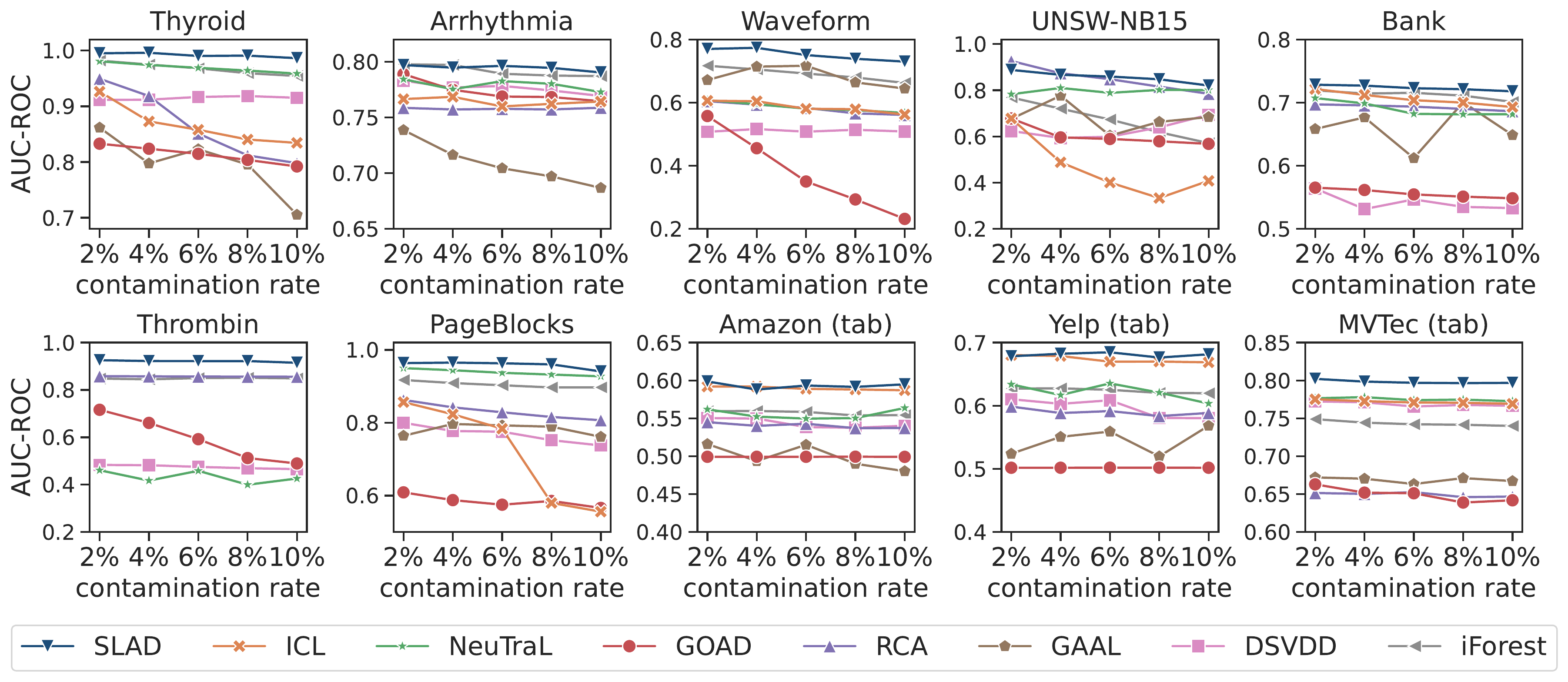}
		\caption{AUC-ROC Performance of SLAD and its competing methods w.r.t. different contamination rate (the percentage of anomalies in training data). 
		}
		\label{fig:contamination}
	\end{figure}
	\subsection{Robustness w.r.t. Different Contamination Levels in Training Data}\label{appendix:contamination}
	
	Recall that we follow the commonly used experimental protocol (half of the normal samples for training and the rest of normal data in addition to all the anomalies for testing) in Section \ref{sec:exp_effectiveness}. However, in real-world applications, training data might be contaminated by unknown anomalies, Thus, we investigate the robustness w.r.t. different contamination ratios in training data. Half of the anomalies are used for testing, and the other half are candidate anomalies to contaminate the training set. As anomalies are rare events, contamination rates are taken from 2\% to 10\%. Following \cite{xu2022deep,pang2019deep}, if the number of candidate anomalies is insufficient to meet the target contamination ratio, we swap 5\% features in two randomly sampled real anomalies to synthesize new anomalies. 
	
	Figure \ref{fig:contamination} depicts the AUC-ROC performance of eight anomaly detectors w.r.t. contamination rate in training data. SLAD shows consistent superiority over its contenders in different scenarios. As these deep anomaly detectors essentially model normal conditions and identify anomalies by measuring the deviation to learned models, contaminated anomalies in training data may mislead the modeling process, leading to the overfitting problem. Generally, detection performance is downgraded with the increase of contamination rate. The increased anomalies may contain diverse behaviors, and thus the performance of some anomaly detectors shows fluctuant trends. In addition, almost all anomaly detectors show stable performance on \textit{Amazon (tab)} and \textit{Yelp (tab)}. These two datasets are challenging, and all the anomaly detectors fail to yield sufficient detection results as reported in Table \ref{tab:effectiveness}. 
	The anomalies in these data are similar to inliers and hard to distinguish since the transferred tabular features may only describe semantic information and not be informative to identify anomalies. 
	Therefore, anomaly detectors might not suffer from the overfitting problem induced by anomaly contamination.

	\subsection{Scalability Test}
	
\begin{figure}[t]
		\centering
		\includegraphics[width=0.66\columnwidth]{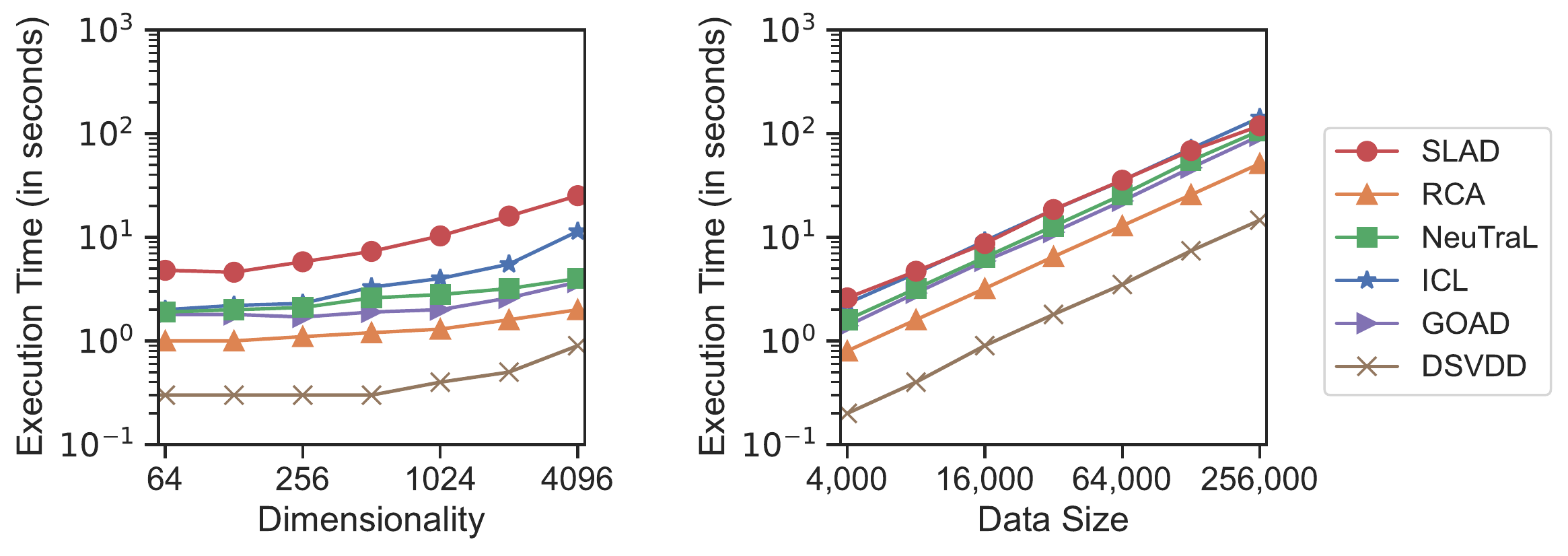}
		\caption{
			Scalability test results. 
		}
		\label{fig:scal}
	\end{figure}

	We further conduct a scalability test to examine the time efficiency of SLAD. A group of synthetic datasets is produced containing 5,000 data instances and varied dimensionality (i.e., \{64, 128, 256, 1,024, 2,048, 4,096\}). We create another suite of data with different data sizes ranging from 4,000 to 256,000 and fixed dimensionality (i.e., 32). These two groups of datasets are used to respectively examine the scalability w.r.t. dimensionality and data size. 
	For the sake of fairness, SLAD is compared with deep anomaly detectors implemented in the PyTorch framework. We report the execution time including the training time of 10 epochs and the inference time.
	Figure \ref{fig:scal} demonstrates the scalability test results.  
	SLAD is less efficient than other anomaly detectors on high-dimensional data. In our implementation, we employ subspaces of the original data, and the neural transformation function $T$ needs to prepare a number of neural layers to handle different subspace lengths. This process may induce relatively heavy overhead. Nevertheless, thanks to the GPU acceleration power, all the anomaly detectors take less than 25 seconds to handle 4,096-dimensional data. In addition, recall that we use a dataset \textit{Thrombin} that contains over one hundred thousand features, SLAD can process this ultra-high-dimensional data in 6 minutes. In terms of the scalability w.r.t. data size, SLAD and contrastive models (ICL, NeuTraL, and GOAD) have comparable time efficiency.
	DSVDD is faster since it maps all the input to a representation and poses a distance-based one-class constraint. However, SLAD and contrastive self-supervised models can lead to superior detection accuracy.

	\begin{figure}[t]
		\centering
		\includegraphics[width=0.85\columnwidth]{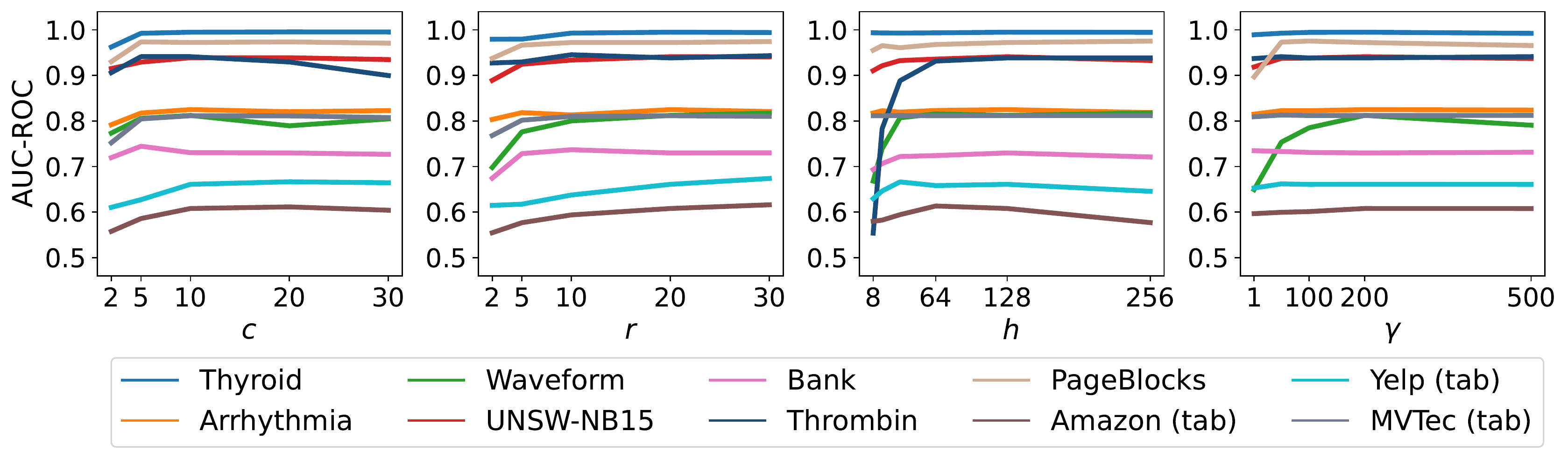}
		\caption{
			Detection performance of SLAD with different hyper-parameter settings. 
		}
		\label{fig:sensitivity}
	\end{figure}

	\subsection{Parameter Sensitivity}
	This experiment investigates the sensitivity of our model to four key hyper-parameters including $c$ (the number of elements in each created data sample of scale learning), $r$ (the factor that adjusts the total size of training samples), $h$ (the dimensionality of representations), and $\gamma$ (the magnification factor in the labeling function $G$), thereby guiding how to set them during practical usages. 
	In each experiment, the tested parameter takes different values, while others are fixed as default. Candidate values for $r$ and $c$ are taken from $\{2,5,10,20,30\}$, the dimensionality $h$ is chosen from $8$ to $256$, and $\gamma$ uses values in $\{1,50,100,200,500\}$.  
	Figure \ref{fig:sensitivity} reports the AUC-ROC performance of SLAD with different settings. In terms of the parameter $c$, the increase of $c$ from 2 to 10 brings observable AUC-ROC gain on most of the datasets, while the performance is clearly downgraded on two datasets when $c$ continually raises. 
	SLAD treats a whole list of transferred data as one training sample, but simultaneously optimizing a very long list of predictions (i.e., a large $c$) may also harm the performance. 
	In terms of $r$, 
	a larger $r$ generally brings better results, which is also true in many similar ensemble scenarios. 
	Based on the above experimental results, we recommend using $n=10$ and $r=20$ in practical usage. 
	Lower dimensionality $h$ cannot contain enough information, which downgrades the detection performance, especially on the high-dimensional dataset \textit{Thrombin}. On the contrary, a large $h$ may be not always feasible. We set $h=128$ by default in SLAD, which is frequently used as representation dimensionality in many deep models. 
	As for the magnification factor $\gamma$, a larger $\gamma$ can enlarge the spacing between different scale values. We use a unified representation dimensionality $h$ for datasets with varied feature numbers, and thus raw scale values are very small and are hard to be differentiated when handling low-dimensional data (their sub-vectors are short). By using a magnification factor, detection performance shows notable improvement on \textit{Waveform} and \textit{PageBlocks}. SLAD performs stably on other datasets w.r.t. $\gamma$.

	\section{Future Directions}
	
	We propose scale learning as a novel self-supervised proxy task for anomaly detection in tabular data. This section notes three future directions that are worth to be explored: (i) SLAD can be generalized to different learning tasks like pre-training representation models on large-scale tabular data, as has been done in \cite{bahri2022scarf,yao2021self,yoon2020vime}. (ii) SLAD can be extended to handle other data types like time series, graph data, or even images by defining appropriate sampling operation, the transformation function, and the labeling function. (iii) The intermediate results in SLAD can be further processed for anomaly interpretation \cite{xu2021beyond}. As empirical errors derived from the loss function indicate the deviation of the target data instance in different feature subspaces, and thus these fine-grained outputs can be further utilized to yield a tailored feature subspace as an interpretation showing the anonymous part of the data instance.


\end{document}